\documentclass[10pt,twocolumn,letterpaper]{article}

\usepackage[pagenumbers]{cvpr} % To force page numbers, e.g. for an arXiv version

\usepackage{algorithm}
\usepackage{algorithmic}
\usepackage[T1]{fontenc}

\definecolor{cvprblue}{rgb}{0.21,0.49,0.74}
\usepackage[pagebackref,breaklinks,colorlinks,allcolors=cvprblue]{hyperref}

\def\paperID{490} % *** Enter the Paper ID here
\def\confName{CVPR}
\def\confYear{2026}

\title{Residual Kolmogorov-Arnold Network for Enhanced Deep Learning}

\author{Ray Congrui Yu \quad Sherry Wu \quad Jiang Gui \\
{\normalsize Dartmouth College, Hanover, New Hampshire}\\
{\tt\small \{ray.yu,sherry.wu.gr,jiang.gui\}@dartmouth.edu}
}

\begin{document}
\maketitle
\begin{abstract}
Despite their immense success, deep convolutional neural networks (CNNs) can be difficult to optimize and costly to train due to hundreds of layers within the network depth. Conventional convolutional operations are fundamentally limited by their linear nature along with fixed activations, where many layers are needed to learn meaningful patterns in data. Because of the sheer size of these networks, this approach is simply computationally inefficient, and poses overfitting or gradient explosion risks, especially in small datasets. As a result, we introduce a ``plug-in" module, called Residual Kolmogorov-Arnold Network (RKAN). Our module is highly compact, so it can be easily added into any stage (level) of traditional deep networks, where it learns to integrate supportive polynomial feature transformations to existing convolutional frameworks. RKAN offers consistent improvements over baseline models in different vision tasks and widely tested benchmarks, accomplishing cutting-edge performance on them.
\end{abstract}    
\section{Introduction}
Convolutional Neural Networks (CNNs) have been widely used in the field of computer vision since the introduction of LeNet and have demonstrated exceptional performance in a variety of image related tasks~\cite{lecun1989backpropagation,lecun2015deep,krizhevsky2012imagenet,khan2020survey}. Followed by the development of modern ConvNet hybrids and Vision Transformers, researchers have been slowly shifting away from convolution-based networks. However, transformer architectures require abundant training samples along with massive computational power, making them more prone to overfitting and generally inferior to traditional CNNs on small datasets~\cite{dosovitskiy2021image} (without pre-training on a much larger dataset). We aim to design a lightweight, add-on module that can reclaim the ``glory" of convolution-based networks by enhancing their ``horizontal" model capacity (especially on datasets of limited data) with distinct, high-quality, yet complementary feature representations.

On the basis of the Kolmogorov-Arnold representation theorem, if $f$ is a multivariate continuous function on a bounded domain, $f$ can be written as a superposition of continuous functions of a single variable~\cite{kolmogorov1957representation}. Compared to MLP (Multi-layer Perceptron), KAN (Kolmogorov-Arnold Network)~\cite{liu2023kolmogorov,liu2024kan} offers a unique perspective in function approximation. Both network architectures retain a fully connected structure, but they handle activations and weights differently. MLP applies fixed activation functions at each neuron, whereas KAN places learnable polynomial-based activation functions along the edges between neurons. As a result, conventional linear weight matrices are entirely replaced by learnable functions, which are parameterized as localized splines or global polynomials. In application, KAN has also excelled in function approximation when compared to standard neural networks~\cite{hou2024kan,yu2024kan,somvanshi2024survey}.

Standard 3$\times$3 convolutional kernels form the backbone of VGG models~\cite{simonyan2014very}, and are later implemented by a wide range of CNN architectures. Each kernel applies a linear combination of each input feature within its receptive field and is typically trained to detect a single type of feature~\cite{lecun1998gradient} (\eg an edge on the feature map). While applying multiple kernels allows the network to learn more diverse features (\eg vertical and horizontal edges), the operation is linear in nature. As a result, even with large numbers of kernels, alongside non-linear activations, models may still struggle to capture more sophisticated spatial dependencies or model curved shapes without relying on additional layers~\cite{luo2016understanding}.

To integrate KAN into the CNN framework, researchers have developed a KAN-based convolution that operates on extracted patches from the input tensor~\cite{bodner2024convolutional}. ``KAN-based kernel" that substitutes each weight in the standard weight matrix with a learnable polynomial-based transformation, in contrast, has multiple ``weights" for each input feature in the pixel space, in which the overall kernel complexity is regulated by the degree of polynomials (\eg d + 1 weights for polynomial of degree d). These weights provide more sophisticated feature detection capabilities that are able to learn non-linear spatial feature interactions, where one KAN convolutional kernel could potentially detect more patterns than multiple traditional kernels.

As a result, as an alternative to overly stacking standard convolutional layers to the network depth, we can use fewer layers with KAN-based kernels to maintain the performance of the model, which proves to be especially useful when the data is scarce. A KAN-based convolutional layer has the potential to match or even exceed the expressive power of multiple standard layers, and considerably reduces the risk of overfitting. Feature representation also becomes more efficient in terms of model parameters and memory.

Our main goal is to strengthen traditional networks by integrating hierarchical, KAN-based convolutional layers as an isolated component alongside a specific stage of the main network branch (\eg where spatial dimension of the feature map changes). Features from each stage usually represent a different level of abstraction, where the final stage retains the most conceptual features. We propose a mechanism, called Residual Kolmogorov-Arnold Network (RKAN), which offers three main benefits to standard CNN architectures. First, compared to the main network path, RKAN blocks can represent feature information with more flexibility using learnable activation functions, and capture specialized patterns that are otherwise unable to be detected by standard convolutions alone. Designed in parallel to the main network stage, the module provides a shorter path for gradients to flow during back-propagation, which reduces the probability of vanishing or exploding gradients~\citep{he2016deep} and simultaneously introduces polynomial transformation to the stage. This not only establishes effective regularization, but can also accelerate the training of much deeper networks. Lastly, RKAN's lightweight, unique stage-level design can be seamlessly employed in existing frameworks (or on top of block-level modules) without modifying the backbone.
\section{Related Work}
\label{sec:related_work}
While various ``plug-in" modules, such as attentions, have been proposed to enhance CNNs, RKAN is fundamentally unique. Our design focuses on the concepts of polynomial transformations and parallel learning in order to address the limitations of deep networks in efficiently capturing highly abstract features, while improving gradient flow.

Unlike standard convolutional kernels that only perform linear operations, KAN-based kernels incorporate learnable polynomial basis functions and directly enable non-linear feature transformations. Consequently, the choice of basis function affects the expressive power of KAN-based neural networks. In the original KAN implementation, B-splines are exceptional in modeling continuous functions~\cite{goldman2002bspline,liu2023kolmogorov}, while providing parameter controls over the shape of the learned function, such as the number of splines applied to the overall function representation or the ``smoothness" (polynomial degree) of the basis function. However, this approach comes with higher computational cost compared to standard convolutions.

Gaussian radial basis functions (RBF)~\cite{li2024fastkan} are used as an approximation for the B-spline basis, which is identified as the primary computational bottleneck in any KAN-based operations. By approximating the B-spline basis (up to a linear transformation), RBF increases the forward speed by three times and maintains very comparable accuracy.

Chebyshev polynomials, calculated recursively, can also be as effective for function representation~\citep{ss2024chebyshev}. Due to their uniform approximation and orthogonal properties over the interval $[-1, 1]$, Chebyshev polynomials are well-suited for modeling smooth functions~\citep{rivlin1974chebyshev,mason2002chebyshev}. They converge quickly, which results in accurate approximations even with low-degree polynomials (\eg 3). By integrating Chebyshev polynomials into KAN kernels, we improve their scalability to handle larger datasets with increased parameter efficiency and less computational demand.

The concept of multi-scale feature representation, first introduced in the Feature Pyramid Network (FPN)~\citep{lin2017fpn}, demonstrates the possibility of aggregating features across network stages through lateral connections. The mechanism has been proven to be successful in object detection and segmentation tasks. While FPN focuses on building feature pyramids for multi-scale detection by merging features of different scales, our module, on the other hand, provides a novel parallel processing unit to expand the ``horizontal" capacity of stages with polynomial transformations.
\section{Residual Kolmogorov-Arnold Network}
\label{sec:overview}
We aim to create a standalone skip connection to enclose an entire network stage. In comparison with conventional skip connections that usually span few layers or blocks, our implementation processes features from a high-resolution stage and integrates them into a low-resolution stage with high-level semantics, bypassing any intermediate blocks in the original network path. This interaction allows low-level features (e.g., shapes) to influence high-level features (e.g., object parts) with a different set of learned transformations (e.g., Chebyshev expansion). For example:
\begin{equation}
y_{\text{s}} = F(x_{\text{s}-1}) \oplus \mathcal{H}(x_{\text{s}-k})
\end{equation}

$y_{\text{s}}$ denotes the aggregated output features of the current stage $s$, while $F(x_{\text{s}-1})$ are features from the previous stage processed by the main network path, $\mathcal{H}(x_{\text{s}-k})$ represents features from $k$ stages back relative to the current stage that are transformed by the residual block (RKAN), $\oplus$ denotes feature aggregation (e.g., addition, gating mechanisms).

The aggregated feature space $\mathcal{V}$ learned by standard and KAN-based convolutions, where  $\sigma$ denotes non-linearity and $T_d(x)$ are polynomials of degree $d$, is defined as:
\begin{equation}
\mathcal{V} = \text{span}\{\sigma(W * x)\} + \text{span}\left\{\sum_{d=0}^D \alpha_d T_d(x)\right\}
\end{equation}

We observe that standard convolutions aggregate their weights before applying any non-linear transformations. This property enforces spatial coherence of the local feature space and excels at detecting correlated neighboring pixel patterns (\eg a straight line and edge). Polynomial-based convolutions learn to transform each pixel independently before aggregation. For example, they can understand the value or ``brightness" of a pixel using a set of polynomial transformations. The mechanisms operate at distinct levels of detail (joint vs. independent) within the same receptive field, in which the combined feature space $\mathcal{V}$ can benefit from both spatial coherence (\eg structure) and pixel-level boundary decisions (\eg light, texture).

In traditional neural network architectures, features are processed sequentially, where low-level details may become ``diluted" and compressed in deeper layers~\cite{liu2018pan}. According to the data processing inequality, for a given input $x$, each sequential transformation $F$ can only maintain or reduce mutual information $I$ compared to the original input~\cite{cover2012elements}:
\begin{equation}
I(x_{s-k}; x_{s-1}) \geq I(x_{s-k}; F(x_{s-1})), \quad k \geq 1
\end{equation}

Alternatively, cross-stage connection in RKAN avoids most information bottlenecks with fewer transformations, and enables feature reuse when low-level features remain relevant for high-level understandings.

Since RKAN provides a direct connection in parallel to the main path, all feature information can selectively skip intermediate layers in the main network, thus creating more stable gradient flow during back-propagation. For example, with respect to loss function $L$, the gradient for parameters $\theta_{s-k}$ in stage $s-k$ is defined by the chain rule:
\begin{equation}
\frac{\partial L}{\partial \theta_{s-k}} = \frac{\partial L}{\partial y_s} \cdot \prod_{j=s-k}^{s-1} \frac{\partial x_{j+1}}{\partial x_j} \cdot \frac{\partial x_{s-k}}{\partial \theta_{s-k}}
\end{equation}

In RKAN, we provide an additional gradient path, which can be crucial for model stability and convergence speed:
\begin{equation}
\frac{\partial L}{\partial \theta_{s-k}} = \frac{\partial L}{\partial y_s} \cdot \left(\prod_{j=s-k}^{s-1} \frac{\partial x_{j+1}}{\partial x_j} + \frac{\partial \mathcal{H}(x_{s-k})}{\partial x_{s-k}}\right) \cdot \frac{\partial x_{s-k}}{\partial \theta_{s-k}}
\end{equation}

\subsection{Overview of RKAN}
\begin{figure}[ht]
    \begin{minipage}{0.22\textwidth}
        \centering
        \includegraphics[width=1\linewidth]{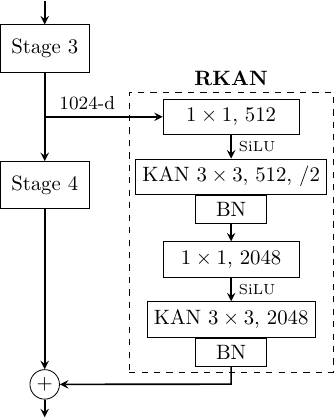}
    \end{minipage}
    \hspace{2mm}
    \begin{minipage}{0.228\textwidth}
        \centering
        \includegraphics[width=1\linewidth]{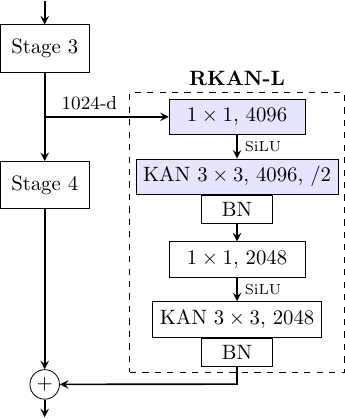}
    \end{minipage}
    % \centering
    % \includegraphics[width = 0.22\textwidth]{figures/resnet_L_architecture.pdf}
    % \hspace{2mm}
    % \includegraphics[width = 0.228\textwidth]{figures/resnet_L_architecture.pdf}
    \caption{RKAN-ResNet-50 (RKANet-50), reduce factor = 2 \textit{(L)}, RKANet-50-4$\times$L, inverse bottleneck expansion multiplier = 4 \textit{(R)}.}
    \label{fig:RKAN}
\end{figure}

RKAN can be implemented into any stage of an established or already enhanced architecture (\eg SENet) and aims to further augment the learning capacity and efficiency of the overall network. The block utilizes kernels parameterized by polynomials and can learn more complex representations through a learnable additive combination of multiple basis polynomial terms of a specified degree.

We implement RKAN and the larger variant RKAN-L to the last stage (level 4) of some of the most widely used CNN frameworks, such as ResNet and DenseNet~\cite{he2016deep,huang2017densely}. This minimizes the interruption of basic feature information processing, reinforces the network to extract most relevant, highly abstract features, and improves gradient flow at the deepest layers where gradient explosion occurs most often, while keeping computational cost low.

Within each RKAN block, several key components are shown in \Cref{fig:RKAN}. The 1$\times$1 bottleneck layers control the number of input and output channels to the RKAN block, while the 3$\times$3 KAN convolutional layers refine and provide non-linearity to the input data. The processed features (final output) from RKAN is then aggregated through summation with the main path output following stage 4.

\subsection{RKAN Block Implementation}
Given an input tensor $X_{\text{in}} \in \mathbb{R}^{B \times C \times H \times W}$, the first 1$\times$1 bottleneck layer is applied to reduce (or expand for the RKAN-L variant) the number of input channels, making feature extraction for subsequent operations more effective. We adopt a depthwise approach for KAN convolution to minimize computational overhead $O(C_{\text{in}})$ vs.\,$O(C_{\text{in}}\times C_{\text{out}})$.

The KAN convolution is performed patch-wise on $X_{\text{in}}$ that consists of $C$ channels of size $H \times W$. For each feature map (channel) $X_{\text{in}}^{(c)} \in \mathbb{R}^{H \times W}$, 3$\times$3 patches are extracted independently. With every individual patch $p = (p_x, p_y)$, $p_x$ and $p_y$ represent the row and column indices of the patch in the feature map. Depending on the base architecture, a stride is used to control how far apart each patch is unfolded and guarantees that the output spatial dimensions of RKAN align with the main network path.

A hyperbolic tangent function $\operatorname{tanh}(p)$ is applied to each patch to ensure the values match the input range of $[-1, 1]$ for Chebyshev polynomials. The normalized output $X_{\text{norm},p}$ then undergoes a Chebyshev expansion:
\begin{equation}
    Y_{\text{chebyshev}, p} = \sum_{i=1}^I \sum_{d=0}^D \alpha_{o,i,d} \cdot T_d(X_{\text{norm},p,i})
    \label{eq:chebyshev_expansion}
\end{equation}

$Y_{\text{chebyshev},p}$ is the output of the Chebyshev expansion for each patch and $\alpha_{o,i,d}$ are learnable weights for the $d^\text{th}$ Chebyshev polynomial of the $o^\text{th}$ output, $i^\text{th}$ input feature, respectively. $I$ is the total number of features for each patch across the spatial dimension (\eg $I$ = 9 for a 3$\times$3 patch). $X_{\text{norm},p,i}$ is the $i^\text{th}$ feature of the $p^\text{th}$ normalized patch. Chebyshev polynomials of degree $d$ are denoted by $T_d$, where $D$ is the maximum degree of the polynomial.

A standard dense layer is applied in parallel to the Chebyshev expansion operation to ensure stability, where a linear transformation is performed to each input patch. The output of the convolution is calculated as the element-wise addition between the polynomial transformed and dense layer output, where the patches are folded back into a tensor of the same spatial dimensions as the strided input, which completes the depthwise KAN convolution process.

Another 1$\times$1 pointwise bottleneck layer is applied to mix channels and match the channel-wise dimension of the last stage in the main network layers. The second 3$\times$3 KAN convolutional layer further processes the feature space to refine high-dimensional spatial feature interactions for smooth integration with main network features. The output of the network before the dense layer (final classification head) is denoted by:
\begin{equation}
    Y_{\text{out}} = F(X_{\text{in}}) + \mathcal{H}(X_{\text{in}})
    \label{eq:output}
\end{equation}

$F(X_{\text{in}})$ is the output of the last stage from the main path and $\mathcal{H}(X_{\text{in}})$ is the output tensor of the RKAN block. The outputs are combined using element-wise addition.
\section{Experiments}
In this experiment, we use widely recognized datasets that consist of different object types and image sizes, such as CIFAR-100, Food-101, Tiny ImageNet, ILSVRC-2012 (ImageNet-1k), MS COCO~\cite{krizhevsky2009learning,bossard14,le2015tiny,deng2009imagenet,lin2014microsoft} to evaluate RKAN's robustness across a broad range of tasks.

\subsection{Training on Image Classification}
\label{sec:training}
To demonstrate the applicability of RKAN with different CNN architectures, we integrate the module at the fourth stage of standard ResNet, ResNeXt, Wide ResNet (WRN), ResNet-D, ResNeSt, Res2Net, ECA-Net, SENet, GCNet, CBAM, PyramidNet, RegNet, DenseNet, SA-Net, SimAM, and ELA~\cite{he2016deep,xie2017aggregated,zagoruyko2016wide,he2019bag,zhang2020resnest,gao2019res2net,wang2020eca,hu2018senet,cao2019gcnet,woo2018cbam,han2017deep,radosavovic2020designing,huang2017densely,zhang2021sa,yang2021simam,xu2024ela}.

For Tiny ImageNet, CIFAR-100, Food-101, all networks are trained from scratch (pre-training) for 200 epochs using stochastic gradient descent (SGD) with a weight decay of 5$\times\text{10}^{-4}$ and Nesterov momentum of 0.9~\cite{sutskever2013importance}. ImageNet-1k is trained using a weight decay of $\text{10}^{-4}$ for 100 epochs. We apply a learning rate scheduler that sets the initial learning rate to 0.005. The learning rate is then incremented by 10 times to 0.05 after 10 linear warm-up epochs, and decreases to $\text{10}^{-5}$, following a cosine annealing schedule~\cite{loshchilov2016sgdr}.

\begin{table}[ht]
\small
\setlength{\tabcolsep}{6.5pt}
\begin{tabular}{@{}l|ccc|cc}
\toprule
\multicolumn{1}{c}{} & \multicolumn{3}{c}{\textbf{+\,RKAN}} & \multicolumn{2}{c}{\textbf{Baseline}}\\
\cmidrule{1-6}
Backbone & \multicolumn{1}{c}{$r$} & \multicolumn{1}{c}{Top-1} & \multicolumn{1}{c|}{img$/$s} & \multicolumn{1}{c}{Top-1} & \multicolumn{1}{c}{img$/$s} \\
\midrule
WRN-101-2~\cite{zagoruyko2016wide} & 1 & \textbf{77.56} & 769 & 75.46 & 881 \\
ResNeXt-101~\cite{xie2017aggregated} & 1 & \textbf{77.48} & 706 & 75.57 & 805 \\
ResNet-152~\cite{he2016deep} & 2 & \textbf{76.82} & 967 & 74.88 & 1,110 \\
ResNet-101 & 2 & \textbf{76.29} & 1,259 & 74.51 & 1,519 \\
ResNeXt-50 & 2 & \textbf{75.41} & 1,443 & 73.56 & 1,779 \\
ResNet-50 & 2 & \textbf{74.43} & 1,686 & 72.85 & 2,159 \\
ResNet-34 & 1 & \textbf{72.03} & 3,012 & 70.96 & 3,412 \\
\midrule
RegNetY-32GF~\cite{radosavovic2020designing} & 2 & \textbf{77.79} & 485 & 75.90 & 541 \\
RegNetY-8GF & 1 & \textbf{77.13} & 890 & 75.58 & 1,025 \\
RegNetY-3.2GF & 2 & \textbf{76.05} & 1,490 & 74.07 & 1,712 \\
DenseNet-161~\cite{huang2017densely} & 2 & \textbf{75.79} & 855 & 74.14 & 947 \\
RegNetX-3.2GF & 1 & \textbf{75.26} & 1,709 & 73.83 & 1,972 \\
DenseNet-201 & 1 & \textbf{75.12} & 1,061 & 73.10 & 1,239 \\
DenseNet-169 & 2 & \textbf{74.88} & 1,355 & 73.55 & 1,548 \\
DenseNet-121 & 2 & \textbf{74.13} & 1,618 & 72.76 & 1,733 \\
RegNetY-800MF & 2 & \textbf{72.19} & 2,801 & 70.43 & 3,003 \\
\bottomrule
\end{tabular}
\caption{Comparison of throughput (img$/$s), Top-1 accuracy (\%) for RKAN-augmented and baseline models on the Tiny ImageNet validation. ($r$) shows RKAN's optimal reduce factor by accuracy.}
\label{tab:acc_tinyimagenet}
\end{table}

\begin{figure*}[ht]
    \centering
    \includegraphics[width=0.325\textwidth]{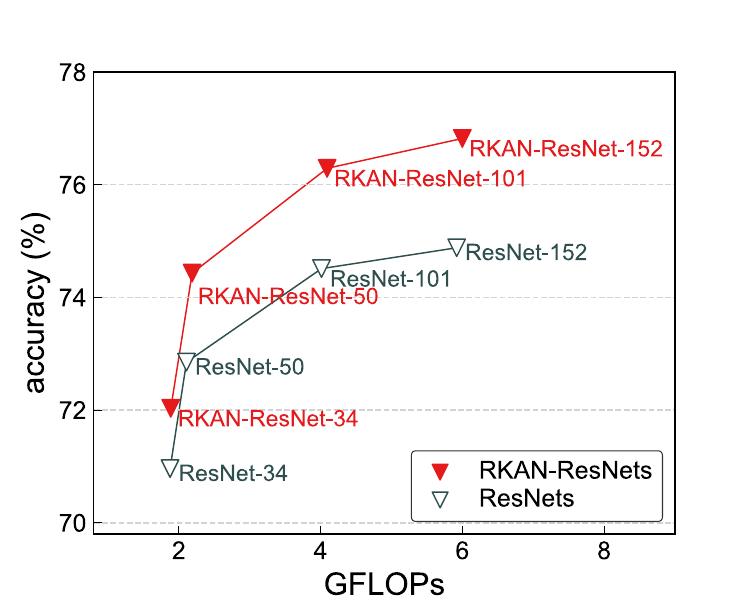}
    \hfill
    \includegraphics[width=0.325\textwidth]{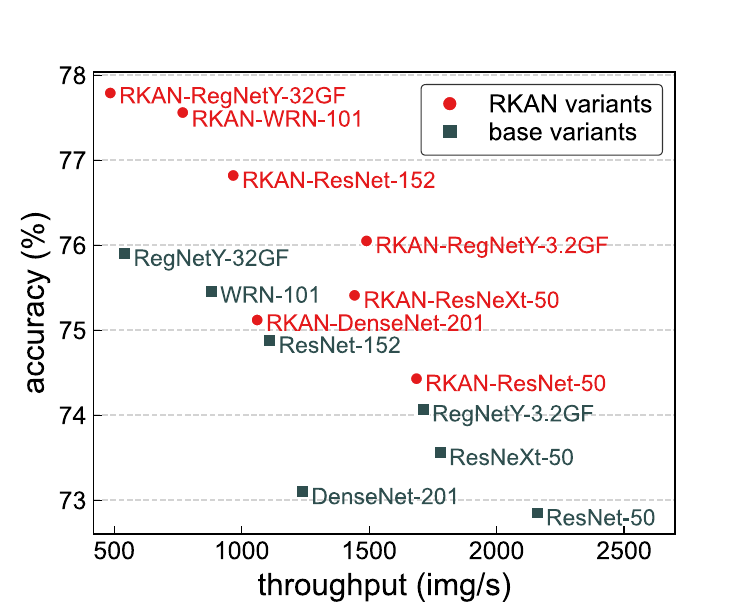}
    \hfill
    \includegraphics[width=0.325\textwidth]{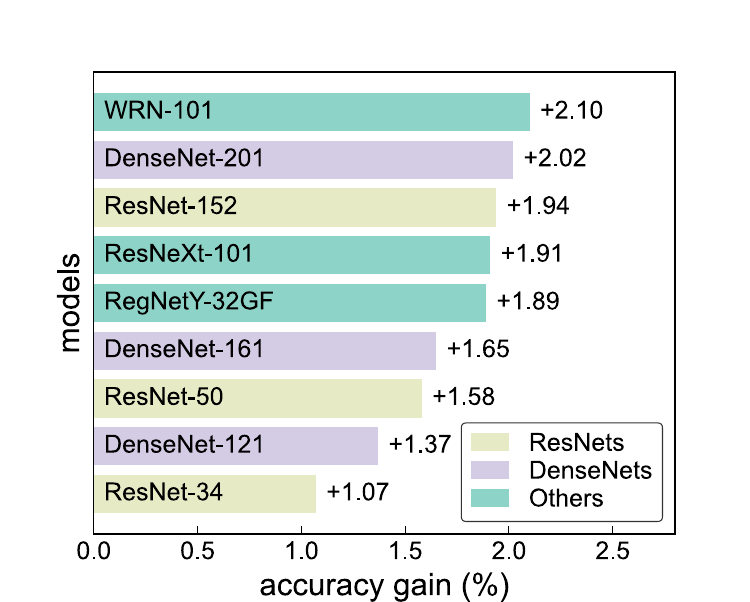}
    \caption{Comparison of RKAN-augmented and baseline model variants in Top-1 accuracy in terms of GigaFLOPs \textit{(L)}, throughput \textit{(Mid)}, and accuracy gain \textit{(R)}, which is calculated as the difference in accuracy between the RKAN-baseline pair.}
    \label{fig:acc_gain_epoch}
\end{figure*}

Across all tests, we use a batch size of 128. According to the linear scaling rule~\citep{goyal2017accurate}, with a large enough batch size, the learning rate should be determined by 0.1$\times\frac{B}{256}$, where $B$ denotes the batch size. We report accuracy using single crop and throughput (img$/$s), defined as $\text{T} = \frac{N}{t}$, where $N$ is the total number of images in the dataset and $t$ is the training time per epoch in seconds. RandAugment, CutMix with 50\% probability ($p$ = 0.5), and MixUp ($\alpha$ = 0.2, $p$ = 0.3) are applied for data augmentation~\cite{cubuk2020randaugment,yun2019cutmix,zhang2018mixup}.

We use the throughput $\text{T}$ (img$/$s) as opposed to FLOPs (floating point operations) or total model parameters as the primary computational metric because the main bottleneck within the implementation of KAN lies in the calculation of basis functions. The additional complexity cannot be fully reflected in the measure of FLOPs as the function ($O(D)$ Chebyshev polynomials) involves multiple recursive steps consisting of several arithmetic operations, as compared to a basic activation, such as ReLU, which performs a simple element-wise computation ($O(1)$).

\subsection{RKAN Parameters}
The RKAN block uses Chebyshev polynomials of degree $\{3, 2\}$ for the first and second KAN layers, respectively. A degree of 3 is chosen mainly for its low computational cost and ability to model non-linear transformations (\eg up to cubic relationships), while maintaining stable gradients. The kernel size for the convolution is fixed at 3$\times$3, while the input is normalized using hyperbolic tangent function $\operatorname{tanh}$. We experiment with 6 channel reduce factors $r$, where $r$ = \{1, 2, 4, 8, 16, 32\}. The reduce factor controls the output channel-wise dimension of the bottleneck layer prior to the first 3$\times$3 layer. RKAN-L adopts inverse bottleneck instead.

\subsection{Results on Tiny ImageNet}
\label{sec:tiny_imagenet}
Tiny ImageNet is a subset of the ILSVRC-2012 ImageNet classification dataset that contains 100,000 images of 200 classes~\cite{le2015tiny}. Each class contains 500 training, 50 validation images. The input of the dataset is 64$\times$64, which causes problems (\eg multiple down-sampling stages, large kernel size) for models originally designed for higher resolution ImageNet data, so we up-scale the training and validation images by 2.5$\times$, to a resolution of 160$\times$160 with bicubic interpolation~\cite{keys1981cubic}. This resolution allows models to retain sufficient spatial details at the last stage, while remaining computationally lightweight compared to 224$\times$224.

We train RKAN-augmented models from scratch using 6 reduce factors ($r$) and report $r$ with the highest accuracy (Top-1) on the validation set. The results are compared with the corresponding baseline models (standard model variants using identical training setup) as shown in \Cref{tab:acc_tinyimagenet}.

\Cref{fig:acc_gain_epoch} shows a trend, where all RKAN-augmented models consistently outperform their standard baselines. Larger variants of the models, such as RKAN-WRN-101 and RKAN-DenseNet-201, are able to improve the Top-1 accuracy from the baseline models by more than 2\%.

Observed frequently in our experiments, a more compact model with the same architectural design, when integrated with the RKAN block, is able to achieve comparable or even higher accuracy than its deeper and wider counterparts. For example, RKAN-ResNet-101 improves by 1\% (on average) over ResNet-152, ResNeXt-101, and WRN-101, which are all ``improved" versions of the original ResNet-101, despite having higher throughput and significantly reduced model complexity (a total of 44.49 million parameters compared to 58.55, 87.15, and 125.25 million, respectively).

We can also observe that scaled-up models have more pronounced performance gains, suggesting that RKAN may scale with size on datasets with limited data and resolution. One reason can be attributed to the fact that larger models tend to overfit more easily on small datasets~\cite{zhang2017understanding}, resulting in under-performance by nature. Once we integrate RKAN into the model, it provides an alternative, but shorter path for feature transformation, which bypasses multiple layers of information flow from the main path and helps regularize the network. This compact feature refinement process helps prevent the model from memorizing specific patterns (\eg when there are more parameters than training examples)~\cite{arpit2017closer}, while focusing on more generalizable features.

\begin{figure*}[ht]
    \centering
    \includegraphics[width=0.325\textwidth]{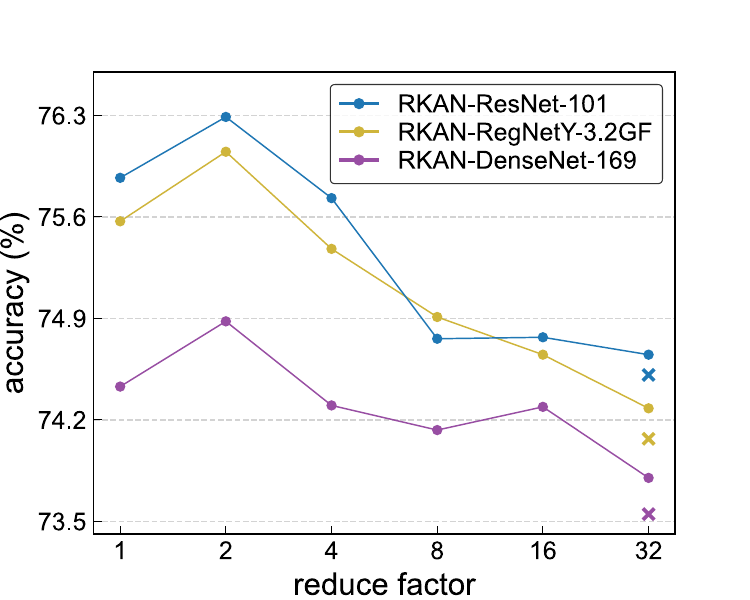}
    \hfill
    \includegraphics[width=0.325\textwidth]{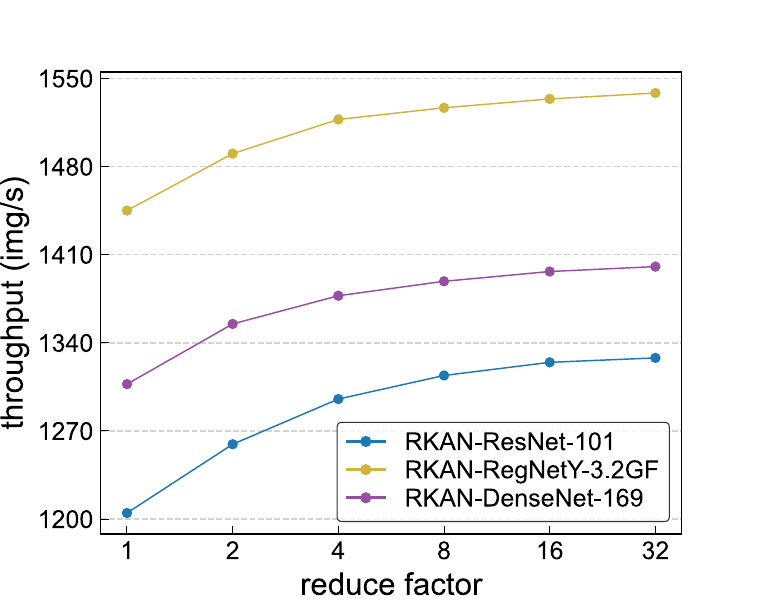}
    \hfill
    \includegraphics[width=0.325\textwidth]{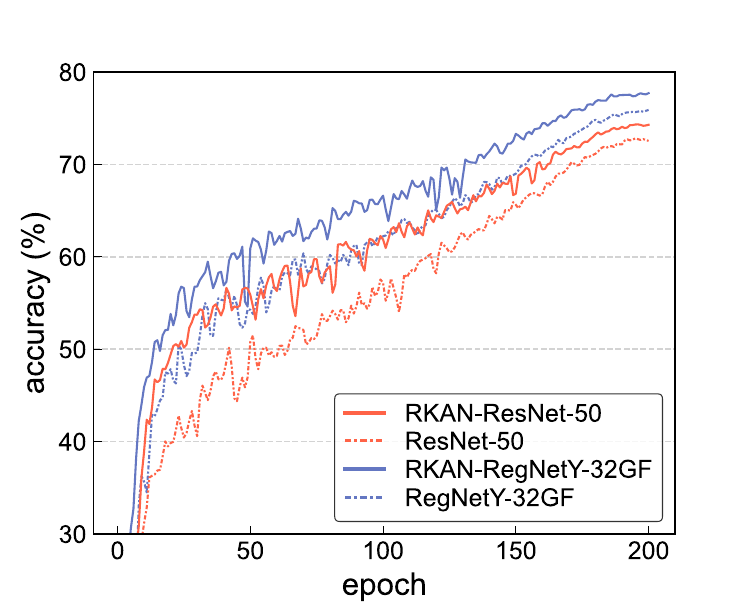}
    \caption{Effect of reduce factor on Top-1 accuracy \textit{(L)}, and throughput \textit{(Mid)} for RKAN-augmented models on the Tiny ImageNet validation. The $\mathbf{x}$ marker indicates the performance of the baseline models. Validation accuracy curves for the training duration \textit{(R)}.}
    \label{fig:rf_acc_time}
\end{figure*}

\paragraph*{Learning Dynamics.} The learning curves presented in \Cref{fig:rf_acc_time} exhibits vastly different convergence rates between RKAN-augmented and baseline models. We observe that the augmented models can achieve higher accuracy than their counterparts from the first few epochs and keep the lead throughout the entire training process. For example, RKAN-ResNet-50 achieves accuracies of 30\%, 50\%, 60\%, and 70\% at epoch 6, 21, 74, 158, respectively, while the standard ResNet-50 only achieves the same accuracy results at epoch 9, 41, 115, 170. The consistent gap in performance indicates that the RKAN module is effective across a wide range of optimization step sizes (e.g., learning rates = 0.05, 0.001, ${10}^{-5}$) and can greatly accelerate model convergence.

\paragraph*{Computational Efficiency.} Models (where $r\:$$\ge$ 2) remain largely efficient with the addition of RKAN as shown in \Cref{fig:rf_acc_time}. For example, RKAN-augmented ResNet-152 or DenseNet-169 ($r$ = 2) only reduces the overall throughput by 12\% (comparable to lightweight attention modules). The overhead becomes even smaller, less than 10\% if $r\:$$\ge$ 8.

To illustrate further, RKAN-augmented RegNetY-3.2GF ($r$ = 2) processes 13\% less images per second compared to baseline RegNetY-3.2GF, but improves the accuracy by almost 2\%. It outperforms the much larger RegNetY-32GF model by 0.15\% (18.83M vs. 142.08M model parameters) with nearly 3$\times$ more throughput. This demonstrates that with the implementation of one RKAN module at the last stage, the network not only achieves better performance, but also becomes more efficient than adding increasing amount of standard convolutional blocks.

\paragraph*{Impact of Reduce Factor.} Reduce factor $r$ controls the bottleneck compression applied to the input channels that enter the first KAN convolutional layer within each RKAN block, which affects both the computational efficiency and capacity of the model. Most modern architectures, such as ResNet, SqueezeNet~\cite{iandola2016squeezenet}, are able to encode important information in a condensed channel space very effectively, and this reduces considerable amount of training overhead without forfeiting much performance.

In \Cref{tab:acc_tinyimagenet}, reduce factors $r$ = \{1, 2\} yield the highest accuracy as small reduce factors are able to preserve more details during the channel reduction process. However, we observe in \Cref{fig:rf_acc_time} that a reduce factor of $r$ = 1 results in significantly lower throughput compared to $r$ = 2, but fails to always yield a higher accuracy in return. For example, RKAN-ResNet-101 ($r$ = 2) obtains an accuracy of 76.29\%, which is 0.42\% higher than the results obtained by $r$ = 1. As we increase the reduce factor $r$ beyond a threshold of 4, performance drops substantially. When the input features are compressed to such an extent, the bottleneck's ability to retain discriminative features during the compression, and more importantly the subsequent expansion process (where the expand factor must double the reduce factor to match the output dimension of the main network stage), diminish as a consequence~\cite{lin2024reducing}. This configuration of $r = 2$ is adopted in the subsequent experiments.

\subsection{Results on CIFAR-100 and Food-101}
\label{sec:cifar_100}
CIFAR-100 (32$\times$32) contains 50,000 training and 10,000 test images across 100 classes (600 samples per class). Food-101 has 101,000 food images evenly distributed across 101 categories, where each category consists of 750 training and 250 test images varying in size. The images are re-scaled to the size of 128$\times$128, 224$\times$224, respectively.

\begin{table}[ht]
\small
\setlength{\tabcolsep}{3.6pt}
\begin{tabular}{@{}l|ccc|ccc}
\toprule
\multicolumn{1}{c}{} & \multicolumn{3}{c}{\textbf{+\,RKAN}} & \multicolumn{3}{c}{\textbf{Baseline (w/o RKAN)}}\\
\cmidrule{1-7}
Backbone & \multicolumn{1}{c}{CV$_{200}$} & \multicolumn{1}{c}{CV$_{100}$} & \multicolumn{1}{c|}{CV$_{50}$} & \multicolumn{1}{c}{CV$_{200}$} & \multicolumn{1}{c}{CV$_{100}$} & \multicolumn{1}{c}{CV$_{50}$} \\
\midrule
ResNeXt-101    & 16.36 & \textbf{3.27} & 1.32 & 18.92 & 4.48 & 1.59 \\
DenseNet-169   & 14.49 & 4.08 & 1.35 & 16.50 & 4.31 & 1.75 \\
RegNetY-8GF    & 15.78 & 3.41 & \textbf{1.29} & 16.88 & 4.21 & 1.48 \\
ECA-Net-101    & \textbf{14.36} & 3.85 & 1.58 & 17.44 & 5.07 & 2.04 \\
\bottomrule
\end{tabular}
\caption{The coefficient of variation (CV) for RKAN-augmented (+\,RKAN) and baseline models on the CIFAR-100 test set. CV is calculated for the last $\{$50, 100, 200$\}$ epochs of each training run.}
\label{tab:cifar_cv}
\end{table}

As detailed in \Cref{tab:acc_cifar}, the addition of RKAN displays consistent performance improvements in both CIFAR-100 and Food-101 when compared to the baseline architectures. We observe on CIFAR-100, deeper variants within the same model family are more prone to overfitting and instability. For example, DenseNet-169 (12.65 million parameters) has more layers than DenseNet-121 (7.06 million parameters), but it is outmatched in terms of accuracy. When augmented with the RKAN module, RKAN-DenseNet-169 is not only able to retain its expected superior performance compared to its smaller variant, but also achieves an accuracy gain of 0.84\% (vs. baseline version). In addition, \Cref{tab:modern_comparison} shows that the RKAN-augmented models achieve state-of-the-art performance on CIFAR-100, against Vision Transformers, modern ConvNets, and also other enhanced architectures. RKAN-ResNet-101 (shortened: RKANet-101) outperforms all other ``enhanced versions" of 101 layers, or models with similar complexity. RKANet-101-6$\times$L improves the result further by 1.95\%, outperforming even PyramidNet-200 by 0.33\%. This showcases our module's ability to regularize the network and improve model capacity, strengthening our theory again that RKAN excels at small-scale datasets.

\begin{table}[ht]
\small
\setlength{\tabcolsep}{3.7pt}
\begin{tabular}{@{}l|cc|cc}
\toprule
\multicolumn{1}{c}{} & \multicolumn{2}{c}{\textbf{CIFAR-100}} & \multicolumn{2}{c}{\textbf{Food-101}}\\
\cmidrule{1-5}
Backbone & \multicolumn{1}{c}{+\,RKAN} & \multicolumn{1}{c|}{Baseline} & \multicolumn{1}{c}{+\,RKAN} & \multicolumn{1}{c}{Baseline} \\
\midrule
ResNeXt-101~\cite{xie2017aggregated} & \textbf{86.15} & 85.28 & \textbf{90.82} & 89.87 \\
ResNet-101-D~\cite{he2019bag} & \textbf{86.07} & 85.09 & \textbf{91.02} & 90.06 \\
Res2Net-101~\cite{gao2019res2net} & \textbf{85.84} & 84.38 & \textbf{90.93} & 89.92 \\
GCNet-101~\cite{cao2019gcnet} & \textbf{85.68} & 84.42 & \textbf{90.78} & 89.98 \\
SA-Net-101~\cite{zhang2021sa} & \textbf{85.54} & 84.40 & \textbf{90.55} & 89.87 \\
ResNeSt-101~\cite{zhang2020resnest} & \textbf{85.52} & 84.47 & \textbf{90.49} & 89.90 \\
SENet-101~\cite{hu2018senet} & \textbf{85.39} & 84.20 & \textbf{90.33} & 89.42 \\
ECA-Net-101~\cite{wang2020eca} & \textbf{85.18} & 84.23 & \textbf{90.59} & 89.81 \\
ResNet-101 & \textbf{\underline{85.12}} & 84.00 & \textbf{\underline{90.09}} & 89.29 \\
\midrule
ResNet-152~\cite{he2016deep} & \textbf{85.40} & 84.63 & \textbf{90.36} & 89.70 \\
ResNeXt-50 & \textbf{85.08} & 84.40 & \textbf{90.00} & 89.20 \\
PyramidNet-200~\cite{han2017deep} & \textbf{86.35} & 85.62 & \textbf{90.61} & 90.05 \\
RegNetY-32GF~\cite{radosavovic2020designing} & \textbf{87.03} & 85.44 & \textbf{91.62} & 90.72 \\
RegNetY-8GF & \textbf{86.11} & 84.77 & \textbf{91.17} & 90.43 \\
DenseNet-201~\cite{huang2017densely} & \textbf{85.35} & 84.28 & \textbf{89.58} & 88.83 \\
DenseNet-169 & \textbf{84.84} & 84.00 & \textbf{89.74} & 89.17 \\
DenseNet-121 & \textbf{84.73} & 84.09 & \textbf{89.43} & 88.98 \\
\bottomrule
\end{tabular}
\caption{Top-1 accuracy (\%) for RKAN-augmented (+\,RKAN) and baseline models on the CIFAR-100 and Food-101 test sets.}
\label{tab:acc_cifar}
\end{table}

On CIFAR-100, the average accuracy gain from RKAN for all tested models is 1.04\%, where larger networks, such as RegNetY-32GF, DenseNet-201, and GCNet-101, achieve gains noticeably above 1\%. Food-101, which contains more and higher resolution images, also follows a similar trend. Despite its reduced tendency to overfit and higher absolute accuracy ceilings on baseline models ($>$\:90\%), Food-101 still achieves an average accuracy improvement of 0.76\%.

We find the implementation of RKAN also stabilizes model performance. The coefficient of variation (CV), is a metric used to measure the stability among test accuracies between epochs over a training run, defined as $\frac{\sigma}{\mu}\times100\%$.

$\sigma$ is the standard deviation, while $\mu$ is the mean value of validation accuracies. A lower CV value suggests that there is less fluctuation with respect to the average accuracy and this typically results in more consistent performance across epochs. In \Cref{tab:acc_cifar}, RKAN-augmented models all retain lower CV over the training run for the last $\{$50, 100, 200$\}$ epochs compared to the baselines, suggesting that RKAN's shortcut block in parallel to the main network stage helps with the network's gradient flow, which in turn smooths the optimization trajectory and accelerates convergence. In practice, the improved stability demonstrates more reliable training dynamics, which can be equally as important as the peak accuracy for model deployment.

\subsection{Results on ImageNet}
\label{sec:imagenet}
ImageNet is made up of over 1.2 million training and 50,000 validation images, consisted of 1,000 classes. Images are resized to 224$\times$224 for training and to 256$\times$256 before center-cropped to a resolution of 224$\times$224 for validation.

\begin{table}[ht]
\small
\setlength{\tabcolsep}{4.7pt}
\begin{tabular}{@{}l|cccc}
\toprule
Model & \multicolumn{1}{c}{Top-1} & \multicolumn{1}{c}{img$/$s} & \multicolumn{1}{c}{FLOPs} & \multicolumn{1}{c}{\#\,Param.} \\
\midrule
MobileNetV4-L$^{224}$~\cite{qin2024mobilenetv4} & 83.24 & 1,230 & 2.17G & 31.16M \\
CoAtNet-1$^{224}$~\cite{dai2021coatnet} & 83.36 & 382 & 8.22G & 41.35M \\
ConvNeXt-S$^{224}$~\cite{liu2022convnet} & 83.52 & 655 & 8.68G & 49.49M \\
EfficientNetV2-M$^{224}$~\cite{tan2021efficientnetv2} & 83.65 & 592 & 5.36G & 52.69M \\
FastViT-MA36$^{224}$~\cite{vasu2023fastvit} & 83.87 & 335 & 6.00G & 42.89M \\
EfficientViT-L2$^{224}$~\cite{liu2023efficientvit} & 85.04 & 650 & 6.95G & 60.83M \\
MaxViT-T$^{224}$~\cite{tu2022maxvit} & 86.40 & 424 & 5.33G & 30.32M \\
\midrule
ResNet-101$^{224}$~\cite{he2016deep} & 85.28 & 741 & 7.86G & 42.71M \\
ResNeXt-101$^{224}$~\cite{xie2017aggregated} & 86.39 & 403 & 16.5G & 86.95M \\
RKANet-101$^{*224}$ & \textbf{\underline{86.81}} & 616 & 8.02G & 44.28M \\
RKANeXt-101$^{*224}$ & \textbf{87.64} & 364 & 16.7G & 88.53M \\
\midrule
ELA-L-101~\cite{xu2024ela} & 84.14 & 906 & 2.60G & 44.53M \\
% SENet-101~\cite{hu2018senet} & 84.20 & 1,887 & 2.58G & 47.45M \\
ECA-Net-101~\cite{wang2020eca} & 84.23 & 1,908 & 2.57G & 42.71M \\
Res2Net-101~\cite{gao2019res2net} & 84.38 & 1,623 & 2.66G & 43.36M \\
SA-Net-101~\cite{zhang2021sa} & 84.40 & 1,111 & 2.57G & 42.71M \\
GCNet-101~\cite{cao2019gcnet} & 84.42 & 1,748 & 2.58G & 47.46M \\
ResNeSt-101~\cite{zhang2020resnest} & 84.47 & 1,524 & 2.57G & 46.31M \\
ReXNet-300~\cite{han2021rethinking} & 84.51 & 1,767 & 1.11G & 31.13M \\
DLA-169~\cite{yu2018deep} & 84.66 & 1,437 & 3.80G & 52.47M \\
PyramidNet-101~\cite{han2017deep} & 84.75 & 1,845 & 2.00G & 26.59M \\
WRN-101-2~\cite{zagoruyko2016wide} & 84.77 & 1,176 & 7.46G & 125.0M \\
DenseNet-161~\cite{huang2017densely} & 84.98 & 1,381 & 2.56G & 26.69M \\
TResNetV2-L~\cite{ridnik2021tresnet} & 85.03 & 1,901 & 2.88G & 44.23M \\
\midrule
RKANet-101$^{*}$ & \textbf{\underline{85.12}} & 1,852 & 2.62G & 44.28M \\
RKANet-E-101$^{*}$ & 85.44 & 1,689 & 2.67G & 44.68M \\
RKANet-101-2$\times$L$^{*}$ & 85.48 & 1,648 & 2.78G & 49.00M \\
RKANet-101-4$\times$L$^{*}$ & 85.66 & 1,412 & 2.97G & 55.30M \\
RKANet-101-6$\times$L$^{*}$ & \textbf{85.95} & 1,210 & 3.17G & 61.60M \\
\bottomrule
\end{tabular}
\caption{Comparison of RKAN-augmented ($^{*}$) and other SOTA models on CIFAR-100. Images are scaled to: 128$^{2}$ (default), 224$^{2}$. (E) denotes RKAN is added to stages \{3, 4\} (details in appendix).}
\label{tab:modern_comparison}
\end{table}

In this experiment, we implement RKAN using RBF (Gaussian radial basis functions) and default Chebyshev polynomials for comparison. We apply three radial basis functions for each RBF-based KAN convolution. In $\frac{11}{12}$ of our results, polynomial-based RKAN block outperforms the alternative with $\sim$300 seconds less training time per epoch.

\begin{table*}[ht]
\small
\setlength{\tabcolsep}{5.3pt}
\begin{tabular}{@{}l|cccc|cccc|cccc}
\toprule
\multicolumn{1}{c}{} & \multicolumn{4}{c}{\textbf{+\,RKAN (Chebyshev)}} & \multicolumn{4}{c}{\textbf{+\,RKAN (RBF)}} & \multicolumn{4}{c}{\textbf{Baseline (w/o RKAN)}} \\
\cmidrule{1-13}
Backbone & \multicolumn{1}{c}{Top-1} & \multicolumn{1}{c}{Top-5} & \multicolumn{1}{c}{img$/$s} & \multicolumn{1}{c|}{\#\,Param.} & \multicolumn{1}{c}{Top-1} & \multicolumn{1}{c}{Top-5} & \multicolumn{1}{c}{img$/$s} & \multicolumn{1}{c|}{\#\,Param.} & \multicolumn{1}{c}{Top-1} & \multicolumn{1}{c}{Top-5} & \multicolumn{1}{c}{img$/$s} & \multicolumn{1}{c}{\#\,Param.} \\
\midrule
ResNet-101~\cite{he2016deep}  & \textbf{80.09} & 94.99 & 687 & 46.13M & 79.95 & 95.08 & 581 & 46.13M & 79.31 & 94.74 & 815 & 44.55M \\
ELA-L-50~\cite{xu2024ela}  & \textbf{78.92} & 94.40 & 505 & 27.98M & 78.86 & 94.46 & 447 & 27.98M & \underline{78.23} & 94.31 & 578$^{\dagger}$ & 26.40M \\
SA-Net-50~\cite{zhang2021sa} & \textbf{78.77} & 94.47 & 483 & 27.14M & 78.69 & 94.34 & 429 & 27.14M & \underline{78.12} & 94.28 & 550$^{\dagger}$ & 25.56M \\
ECA-Net-50~\cite{wang2020eca} & 78.69 & 94.44 & 784 & 27.14M & \textbf{78.73} & 94.55 & 673 & 27.14M & \underline{77.90} & 94.14 & 972 & 25.56M \\
SENet-50~\cite{hu2018senet}  & \textbf{78.66} & 94.31 & 779 & 29.65M & 78.44 & 94.41 & 668 & 29.65M & \underline{77.68} & 94.01 & 965 & 28.07M \\
CBAM-50~\cite{woo2018cbam}  & \textbf{78.45} & 94.28 & 589 & 29.65M & 78.38 & 94.32 & 507 & 29.65M & \underline{77.81} & 93.97 & 694 & 28.07M \\
SimAM-50~\cite{yang2021simam} & \textbf{78.24} & 94.08 & 396 & 27.14M & 78.12 & 94.15 & 359 & 27.14M & \underline{77.59} & 93.85 & 438$^{\dagger}$ & 25.56M \\
ResNet-50       & \textbf{\underline{78.02}} & 93.92 & 943 & 27.14M & 77.94 & 94.00 & 766 & 27.14M & 77.15 & 93.67 & 1,216 & 25.56M \\
\midrule
RegNetY-3.2GF~\cite{radosavovic2020designing} & \textbf{79.62} & 94.82 & 859 & 20.04M & 79.58 & 94.84 & 742 & 20.04M & 79.03 & 94.64 & 998 & 19.44M \\
DenseNet-201~\cite{huang2017densely} & \textbf{79.02} & 94.63 & 615 & 21.28M & 78.89 & 94.53 & 533 & 21.28M & 78.41 & 94.25 & 701 & 20.01M \\
DenseNet-169    & \textbf{78.00} & 94.08 & 770 & 14.89M & 77.98 & 94.03 & 662 & 14.89M & 77.25 & 93.75 & 843 & 14.15M \\
DenseNet-121    & \textbf{76.34} & 93.24 & 947 & 8.37M & 76.25 & 93.23 & 840 & 8.37M & 75.05 & 92.48 & 1,054 & 7.98M \\
\bottomrule
\end{tabular}
\caption{Top-1 accuracy, runtime throughput (img$/$s), model parameters compared between RKAN-augmented (Chebyshev polynomials vs. RBFs) and baseline models on ImageNet-1k. Memory-bound spatial attention methods ($>$\:100\% training overhead) are marked ($^{\dagger}$).}
\label{tab:acc_imagenet}
\end{table*}

\begin{table}[ht]
\small
\setlength{\tabcolsep}{5pt}
\begin{tabular}{@{}l|ccccc}
\toprule
Backbone & \multicolumn{1}{c}{AP$^{\text{bbox}}$} & \multicolumn{1}{c}{AP$^{\text{bbox}}_{50}$} & \multicolumn{1}{c}{AP$^{\text{mask}}$} & \multicolumn{1}{c}{AP$^{\text{mask}}_{50}$} & \multicolumn{1}{c}{FPS} \\
\midrule
RKAN-SENet-50 & \textbf{36.35} & \textbf{54.48} & \textbf{32.37} & \textbf{51.64} & 82.4 \\
RKANet-50-2$\times$L & 36.13 & 54.30 & 32.29 & 51.38  & 97.2 \\
RKANet-50  & 35.92 & 54.21 & 32.16 & 51.20  & 105.5 \\
\midrule
SA-Net-50~\cite{zhang2021sa} & 36.20 & 54.39 & 32.20 & 51.34 & 79.9 \\
ECA-Net-50~\cite{wang2020eca}  & 35.98 & 54.04 & 32.09 & 51.26  & 93.2 \\
SENet-50~\cite{hu2018senet}  & 35.94 & 54.10 & 32.14 & 51.13 & 90.0 \\
ResNet-50~\cite{he2016deep}  & 35.59 & 53.58 & 31.94 & 50.79 & 118.2 \\
\bottomrule
\end{tabular}
\caption{Comparison of AP$_{50}$ (IoU = 0.5), AP (IoU = 0.5:0.95), FPS (forward pass inference speed) between RKAN and different enhanced backbones in Mask R-CNN on MS COCO 2017.}
\label{tab:coco}
\end{table}

Our results in \Cref{tab:acc_imagenet} and \Cref{tab:imagenet_ablation} show that RKAN, and especially RKAN-L can also be effective on large datasets, such as ImageNet-1k. Since the original baseline models are specifically well-optimized on the dataset and less prone to overfitting, improvements are less pronounced compared to CIFAR-100, but RKANet-50 and RKANet-50-2$\times$L still outperform other channel and spatial attention modules of similar computational overhead and can also be integrated alongside them to further improve accuracy. This makes the combination of two enhancement methods, attention-based feature recalibration mechanism operating on blocks and our polynomial-based parallel feature learning mechanism across stages, possible all in a single model. While RKAN leads to more pronounced gains for larger models on small datasets, lightweight models perform better on ImageNet. For example, RKAN-DenseNet-121 improves the baseline accuracy by as much as 1.29\%, but RKAN-DenseNet-201 only achieves an accuracy gain of 0.61\%. This shows that when training data is abundant, RKAN behaves more like a ``feature enhancement" to the main path as opposed to a regularization operation. Deeper models have significantly higher accuracy ceilings and might be close to their optimal architectural capacity, while smaller models (less capacity constraints) are more probable to benefit from RKAN.

\subsection{Object Detection and Instance Segmentation}
MS COCO 2017 consists of 80 object categories with 118,287 training and 5,000 validation samples. All images are resized to 640 pixels on the shorter side.

We apply Mask R-CNN for object detection and instance segmentation~\cite{he2017mask} on MS COCO 2017. We then use our proposed RKANet-50 to compare with other enhancement methods (\eg ECA-Net). Models are trained for 12 epochs using pre-trained weights from ImageNet-1k. We apply SGD with a weight decay of $\text{10}^{-4}$, batch size of 8, and base learning rate of 0.01 that decays by a factor of 10 at epoch 9 and 12. For data augmentation, we apply basic horizontal flipping ($p$ = 0.5), random scaling ($\leq$\:10\%), color jittering.

\begin{table}[ht]
\small
\setlength{\tabcolsep}{4.7pt}
\begin{tabular}{@{}l|ccccc}
\toprule
RKAN Variant & \multicolumn{1}{c}{Top-1} & \multicolumn{1}{c}{$\Delta$} & \multicolumn{1}{c}{img$/$s} & \multicolumn{1}{c}{FLOPs} & \multicolumn{1}{c}{\#\,Param.} \\
\midrule
RKANet-50-6$\times$L & \textbf{78.91} & +1.76 & 500 & 5.99G & 44.45M \\
RKANet-50-4$\times$L & 78.80 & +1.65 & 632 & 5.37G & 38.15M \\
RKANet-50-2$\times$L & 78.65 & +1.50 & 780 & 4.75G & 31.86M \\
RKANet-50 & 78.02 & +0.87 & 943 & 4.29G & 27.14M \\
\bottomrule
\end{tabular}
\caption{Comparison of RKAN and RKAN-L (2$\times$, 4$\times$, 6$\times$) on ImageNet-1k. $\Delta$ denotes improvement over baseline ResNet-50.}
\label{tab:imagenet_ablation}
\end{table}

In \Cref{tab:coco}, all RKAN variants improve upon ResNet-50 with minimal overhead and outperform in most metrics with faster inference speed in both object detection and instance segmentation, when compared to enhancement methods of similar computational cost. This shows that our backbone has strong generalizability (despite being integrated only at stage 4) and can be easily transferred to downstream tasks.

\section{Conclusion}
In this paper, we propose a novel ``plug-in" network module, called Residual Kolmogorov-Arnold Network (RKAN). It is integrated in parallel to each stage of the main network structure and seeks to complement standard convolutions or even attentions with learnable polynomial transformations.

RKAN shows strong performance on different tasks, architectures, datasets, but it particularly excels on small datasets due to its simple, compact design, and efficiency in parameter usage, where the implementation of one RKAN module can exceed the performance of multiple standard convolutional blocks. Combined with polynomials, RKAN is highly efficient in both forward and backward speed when compared to the original B-spline approach, in which the extreme computational demand makes training improbable in real-world scenarios. In addition, since RKAN provides an alternative path for gradient flow, we observe improved stability and accelerated model convergence.

\appendix
\section{Implementation Details}
All models are trained and evaluated on Nvidia RTX 4090 GPU. Implementation details on training modern ConvNets, Vision Transformers on CIFAR-100 are shown in \Cref{tab:training_details}. Larger variants (\eg ConvNeXt-L, CoAtNet-3, MaxViT-B) usually underperform against smaller variants due to the lack of inductive bias and potential overfitting. Other CNNs trained in 128$\times$128 use the setup described in \Cref{sec:training}.

\begin{table}[ht]
\small
\setlength{\tabcolsep}{3.9pt}
\begin{tabular}{@{}l|ccc}
\toprule
Model & \multicolumn{1}{c}{\fontsize{8.5pt}{9pt}\selectfont Optimizer} & \multicolumn{1}{c}{\fontsize{8.5pt}{9pt}\selectfont Weight decay} & \multicolumn{1}{c}{\fontsize{8.5pt}{9pt}\selectfont Learning rate} \\
\cmidrule{1-4}
ConvNeXt-S~\cite{liu2022convnet} & AdamW & 0.05 & 0.004 \\
MaxViT-T~\cite{tu2022maxvit} & AdamW & 0.05 & 0.003 \\
EfficientViT-L2~\cite{tan2021efficientnetv2} & AdamW & 0.05 & 0.001 \\
FastViT-MA36~\cite{vasu2023fastvit} & AdamW & 0.05 & 0.001 \\
CoAtNet-1~\cite{dai2021coatnet} & AdamW & 0.1 & 0.001 \\
MobileNetV4-L~\cite{qin2024mobilenetv4} & SGD & 1e-4 & 0.05 \\
EfficientNetV2-M~\cite{tan2021efficientnetv2} & SGD & 1e-4 & 0.05 \\
\midrule
Baseline & SGD & 5e-4 & 0.05 \\
\bottomrule
\end{tabular}
\caption{Implementation details of different architectures trained on CIFAR-100 with random initialization. Models are tested with weight decay values of $[0.025, 0.2]$, optimized for Top-1 accuracy.}
\label{tab:training_details}
\end{table}

\section{Inspiration}
Wide ResNet~\cite{zagoruyko2016wide} has demonstrated the potential of wider networks. For example, WRN-50-2 has a reduction ratio of 2 in the bottleneck layers (128, 256, 512, 1024), whereas ResNet-101 retains a reduction ratio of 4. In comparison, WRN-50-2 outperforms the latter with half the network depth in most tasks, such as CIFAR-100 and ImageNet-1k, showing that ultra-deep network is not the only solution.

We wonder what would happen if networks are scaled horizontally with ``quality" instead of ``quantity". Kolmogorov-Arnold representation theorem states that any continuous multivariate function $f: [0,1]^n \to \mathbb{R}$ can be expressed as:
\begin{equation}
f(x_1, \ldots, x_n) = \sum_{q=0}^{2n} \Phi_q \left(\sum_{p=1}^{n} \phi_{q,p}(x_p)\right)
\end{equation}

$\phi_{q,p}: [0,1] \to \mathbb{R}$, and $\Phi_q: \mathbb{R} \to \mathbb{R}$ are 
continuous univariate functions. The theorem demonstrates that even highly complex functions can be decomposed into simpler, learnable univariate functions and raises the fundamental question of ``more parameters or enriched per parameter transformations". Instead of adding width (channels), we enhance network width with polynomial transformations. With various experiments and empirical evidences, RKAN proves to be valuable, providing more expressive features in return. The cross-stage connection also smooths gradient flow to prevent gradient vanishing or explosion, and offers natural regularization on datasets with limited data.

\begin{table}[htbp]
\centering
\small
\setlength{\tabcolsep}{4.8pt}
\begin{tabular}{@{}l|cc|cc}
\toprule
\multicolumn{1}{c}{} & \multicolumn{2}{c}{\textbf{+\,RKAN}} & \multicolumn{2}{c}{\textbf{Baseline}} \\
\cmidrule{1-5}
Backbone & \multicolumn{1}{c}{FLOPs} & \multicolumn{1}{c|}{\#\,Param.} & \multicolumn{1}{c}{FLOPs} & \multicolumn{1}{c}{\#\,Param.} \\
\midrule
WRN-101~\cite{zagoruyko2016wide} & 11.81G & 128.40M & 11.65G & 125.25M \\
ResNeXt-101~\cite{xie2017aggregated} & 8.60G & 90.30M & 8.44G & 87.15M \\
ResNeXt-50  & 2.27G & 24.97M & 2.19G & 23.39M \\
ResNet-152~\cite{he2016deep}  & 6.00G & 60.13M & 5.92G & 58.55M \\
ResNet-101  & 4.09G & 44.49M & 4.01G & 42.91M \\
ResNet-34   & 1.89G & 21.59M & 1.88G & 21.39M \\
\midrule
RegNetY-32GF~\cite{radosavovic2020designing}  & 16.70G & 145.64M & 16.54G & 142.08M \\
RegNetY-8GF   & 4.49G & 40.38M & 4.37G & 37.77M \\
RegNetX-3.2GF & 1.67G & 15.11M & 1.64G & 14.49M \\
DenseNet-161~\cite{huang2017densely}  & 4.05G & 28.64M & 4.00G & 26.91M \\
DenseNet-201  & 2.30G & 21.01M & 2.24G & 18.48M \\
DenseNet-121  & 1.49G & 7.55M & 1.48G & 7.16M \\
\bottomrule
\end{tabular}
\caption{Comparison of FLOPs, model parameters (in millions) between RKAN-augmented and baselines on Tiny ImageNet.}
\label{tab:flops_tinyimagenet}
\end{table}

\begin{figure}[htbp]
    \centering
    \hspace{-0.15cm}
    \includegraphics[width=0.98\linewidth]{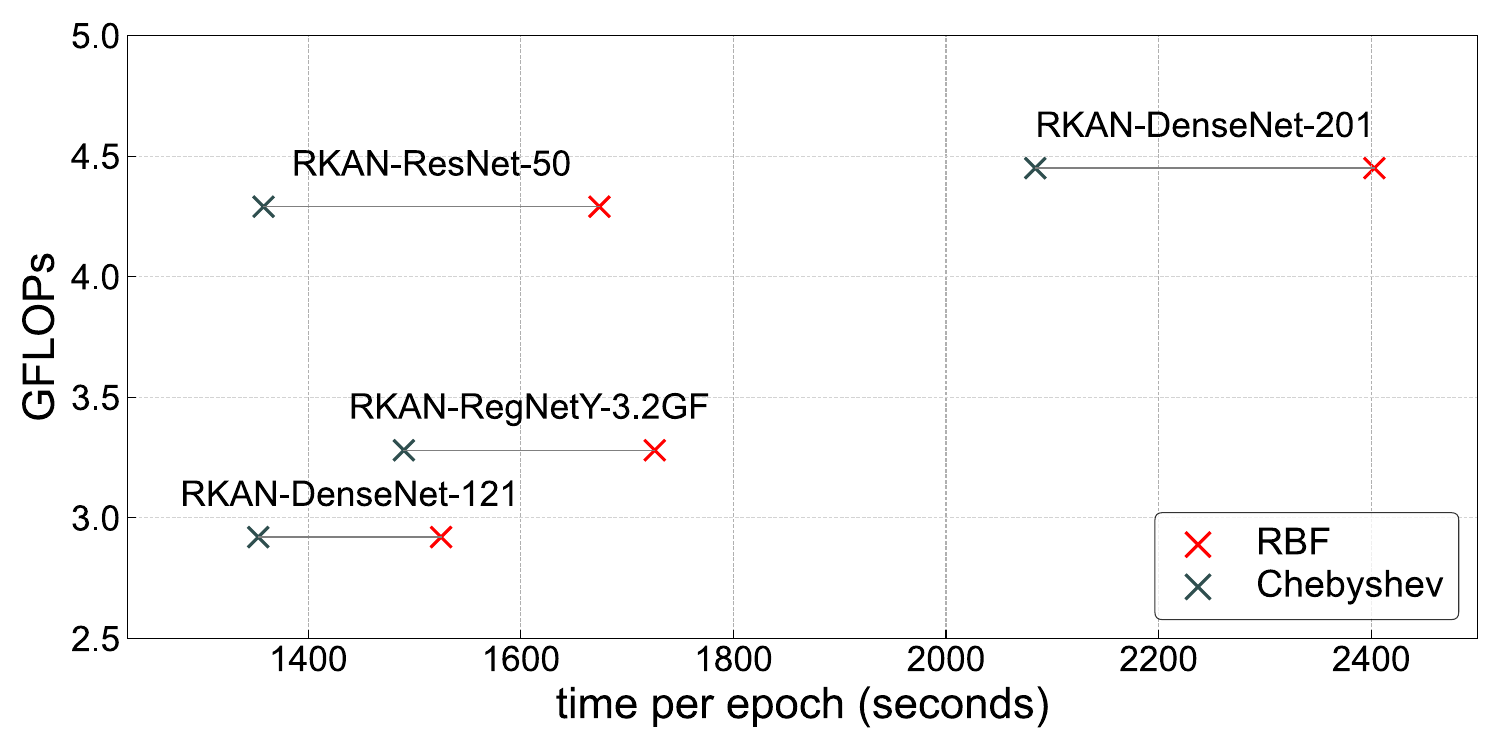}
    \caption{FLOPs and throughput compared between RBF-based and Chebyshev polynomial-based RKAN-augmented models on the ILSVRC-2012 ImageNet dataset~\cite{deng2009imagenet} of resolution 224$\times$224.}
    \label{fig:throughput_flops}
\end{figure}

\section{Computational Metrics}
The total number of model parameters for our KAN-based convolutional layer is primarily determined by the degree ($d$) of Chebyshev polynomials in comparison to a standard depthwise separable convolutional layer, expressed as:
\begin{equation}
KAN_{\text{p, DWC}} = C_{in} \times k_h \times k_w \times (d + 1)
\end{equation}

$C_{in}$ denotes the input channels, while $k_h \times k_w$ is the kernel size (height × width). As shown in \Cref{tab:flops_tinyimagenet}, model parameters for all RKAN-augmented ResNet architectures that have the bottleneck structure ($C_{in} = 1024$, $C_{out} = 2048$) are increased by a fixed amount of 1.58 million. For example, RKANet-152 has 2.7\% more model parameters and 1.4\% more FLOPs than baseline ResNet-152. However, RKANet-152 reduces the throughput by 12.9\% compared to the baseline, which is hardly reflected in the calculation of either metrics above.

\begin{figure*}[ht]
    \centering
    \includegraphics{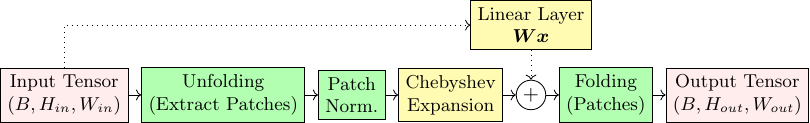}
    \caption{KAN-based convolutional layer implemented using Chebyshev polynomials, applied depthwise (independently to each channel).}
    \label{fig:kan_conv}
\end{figure*}

\begin{center}
    \begin{minipage}{0.47\textwidth}
    \centering
    \vspace{-0.25cm}
    \begin{algorithm}[H]
    \caption{KAN Convolution Process} \label{alg:algorithm}
    \begin{algorithmic}[1]
    \REQUIRE Input tensor $\mathbf{X} \in \mathbb{R}^{B \times C \times H \times W}$, kernel size $k$, stride $s$, padding $p$
    \REQUIRE Chebyshev degree $D$, weights $\alpha_{o,i,d}$
    \ENSURE Output feature maps $\mathbf{Y}$
    \STATE Extract patches from $\mathbf{X}$ using unfold operation
    \STATE Reshape patches for processing
    \STATE Normalize input patches to the range of $[-1, 1]$: $\mathbf{X}_{\text{norm}} \leftarrow \tanh(\text{patches})$
    \STATE Compute Chebyshev polynomial basis functions:
           $T_0 \leftarrow \mathbf{1}$, $T_1 \leftarrow \mathbf{X}_{\text{norm}}$
    \FOR{$d = 2$ to $D$}
        \STATE $T_d \leftarrow 2 \cdot \mathbf{X}_{\text{norm}} \cdot T_{d-1} - T_{d-2}$
    \ENDFOR
    \STATE $\mathbf{T} \leftarrow $ Stack($[T_0, T_1, \ldots, T_D]$)
    \STATE $\mathbf{Y}_{\text{chebyshev}} \leftarrow \sum_{i=1}^{I} \sum_{d=0}^{D} \alpha_{o,i,d} \cdot T_d(\mathbf{X}_{\text{norm},i})$
    \STATE Reshape result to output tensor format
    \RETURN $\mathbf{Y}$
    \end{algorithmic}
    \end{algorithm}
    \end{minipage}
\end{center}

\Cref{fig:throughput_flops} compares multiple RKAN-augmented models based on RBFs, or Chebyshev polynomials by FLOPs and training time. We observe that although both functions achieve almost identical FLOPs (measured to a precision of 100,000), RBF-based RKAN creates a fixed overhead within the implementation of basis function operations. In Gaussian RBF, the exponential calculations can become computationally expensive when compared to Chebyshev polynomials, which are evaluated only using a recurrence relation that performs basic multiplications and additions.

As a result, since the higher computational complexity in evaluating basis functions cannot be properly measured in the calculation of model parameters and FLOPs, along with potential hardware optimization problems (memory-bound operations), such as less efficient memory access patterns, it is more objective to use the throughput as the primary measurement standard across our experiments.

\begin{figure*}[ht]
    \centering
    \scalebox{0.75}{{\includegraphics{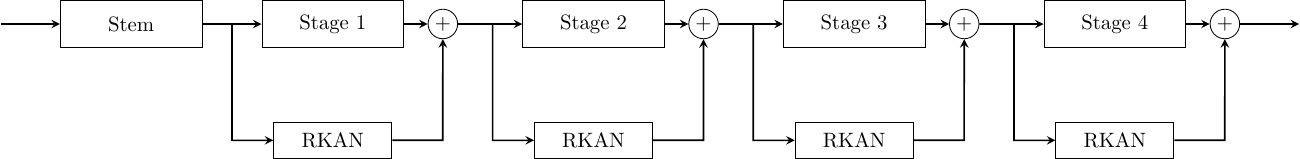}}}
    \caption{RKAN blocks integrated at multiple network stages of the ResNet architecture.}
    \label{fig:rkan_blocks}
\end{figure*}

\section{Details on Depthwise KAN Convolution}
We demonstrate the process of how features are processed within a single KAN convolutional layer, shown in \Cref{fig:kan_conv}. The input tensor should match the output spatial dimension ($H_{in} \times W_{in}$) of the first 1$\times$1 bottleneck layer while the output tensor needs to match the input spatial dimension ($H_{out} \times W_{out}$) of the second bottleneck layer and the main network path output. The channel-wise dimension remains unchanged (due to depthwise convolution) and is regulated by the bottleneck layers for efficiency instead. \Cref{alg:algorithm} provides a more detailed description of the operation.

The polynomial-based KAN layer, represented by the ``Chebyshev Expansion" in \Cref{fig:kan_conv} is primarily inspired by the original spline-based KAN implementation where the activation function $\phi(x)$ is a linear combination of the SiLU activation $\mathrm{silu}(x)$ and the spline function~\cite{liu2023kolmogorov}, in which $N$ denotes the number of splines, $B_i(x)$ are the B-spline basis functions, and $c_i$ are the trainable coefficients:
\begin{equation}
\phi(x) = w_b \, \mathrm{silu}(x) + \sum_{i=1}^{N} c_i B_i(x)
\end{equation}

we have adjusted the spline function to use Chebyshev polynomials instead and also removed the SiLU activation from the skip connection, $w_b \, \mathrm{silu}(x)$, in order to improve model stability, especially when dealing with high-degree polynomials. The activation function $\phi(x)$ is denoted by:
\begin{equation}
\phi(x) = w_b \, x + \sum_{d=0}^{D} \alpha_d T_d(x)
\end{equation}

$D$ represents the maximum degree of the Chebyshev polynomials, while $T_d(x)$ are the polynomials of degree d, and $\alpha_d$ are the trainable coefficients. $T_d(x)$ is defined by the recurrence relation below that makes the calculation of high-degree orthogonal polynomials more efficient:
\begin{equation}
\begin{gathered}
T_0(x) = 1, \quad T_1(x) = x, \\
T_d(x) = 2x\, T_{d-1}(x) - T_{d-2}(x) \quad \text{for } d \geq 2
\end{gathered}
\end{equation}

\section{RKAN Integration at Multiple Stages}
We have implemented the RKAN block into the fourth stage of different baseline models in our previous experiments and observe consistent performance improvements. RKAN can be similarly integrated into other stages of the network as presented in \Cref{fig:rkan_blocks}. However, since the previous stages process feature maps that retain larger spatial dimensions, which may significantly increase the training duration, we remove the second KAN convolutional layer to reduce the extra computational demand, while still preserving the key polynomial transformation shortcut. It should be noted that the integration only works with the standard RKAN variant.

\begin{table}[ht]
\centering
\small
\setlength{\tabcolsep}{4pt}
\begin{tabular}{@{}l|cc|cc|cc}
\toprule
\multicolumn{1}{c}{} & \multicolumn{2}{c}{\textbf{s = \{2, 3, 4\}}} & \multicolumn{2}{c}{\textbf{s = \{3, 4\}}} & \multicolumn{2}{c}{\textbf{s = \{4\}}} \\
\cmidrule{1-7}
Model & \multicolumn{1}{c}{Top-1} & \multicolumn{1}{c|}{img$/$s} & \multicolumn{1}{c}{Top-1} & \multicolumn{1}{c|}{img$/$s} & \multicolumn{1}{c}{Top-1} & \multicolumn{1}{c}{img$/$s} \\
\midrule
RKANeXt-101    & 86.13 & 928 & \textbf{86.51} & 1,044 & 86.15 & 1,119 \\
RKANet-152     & 85.56 & 1,166 & \textbf{86.07} & 1,370 & 85.40 & 1,475 \\
RKANet-101     & 85.15 & 1,420 & \textbf{85.44} & 1,689 & 85.12 & 1,852 \\
RKANeXt-50     & 85.10 & 1,695 & \textbf{85.23} & 2,083 & 85.08 & 2,336 \\
\bottomrule
\end{tabular}
\caption{Comparison of throughput (img$/$s) and Top-1 accuracy (\%) between models with RKAN integrated at different stages on CIFAR-100. (\textbf{s}) includes all stages where RKAN is implemented.}
\label{tab:multiple_rkan}
\end{table}

The RKAN block is integrated with 3 configurations, in which we have tested on CIFAR-100: s = \{2, 3, 4\} at stages 2, 3, and 4; s = \{3, 4\} at stages 3 and 4; s = \{4\} at stage 4 only. In \Cref{tab:multiple_rkan}, we observe s = \{3, 4\} consistently outperforms other configurations, including s = \{2, 3, 4\}, where the RKAN block is additionally incorporated in the second stage. Furthermore, the average throughput for all tested models only decreases by $\sim$8\% compared to s = \{4\}. This suggests that stages with more complex and abstract features benefit most from RKAN. In contrast, low-level features may not require such non-linear transformations and this could result in overfitting as a consequence. The network may need to establish certain fundamental features before RKAN is implemented, while adding the module at an earlier stage (\eg first stage) could disrupt this carefully optimized learning process. We use (E) to denote extended models with s = \{3, 4\} configuration (\eg RKANet-E-101).

\begin{figure}[ht]
    \begin{minipage}{0.22\textwidth}
        \centering
        \includegraphics[width=1\linewidth]{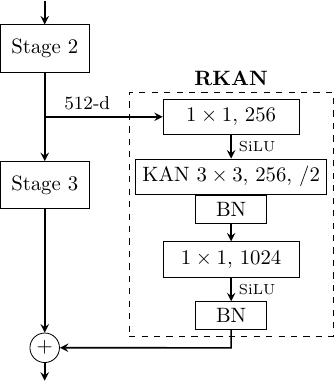}
    \end{minipage}
    \hspace{2mm}
    \begin{minipage}{0.22\textwidth}
        \centering
        \includegraphics[width=1\linewidth]{figures/resnet_architecture.pdf}
    \end{minipage}
    % \centering
    % \includegraphics[width = 0.22\textwidth]{figures/resnet_architecture_stage3.pdf}
    % \hspace{2mm}
    % \includegraphics[width = 0.22\textwidth]{figures/resnet_architecture.pdf}
    \caption{RKAN implemented at stage 3 of ResNet-101 with one 3$\times$3 layer after channel reduction \textit{(L)}, and standard RKAN with two 3$\times$3 layers implemented at stage 4 of the network \textit{(R)}.}
    \label{fig:RKAN_multi}
\end{figure}

In \Cref{fig:RKAN_multi}, the left diagram illustrates the RKAN block at stage 3 of RKANet-101, which excludes the second KAN convolutional layer operating on full channels, subsequent to the second bottleneck layer. For example, in RKANet, other components remain unchanged as the module takes the output of stage 2 (512 channels) and ``compresses" the number of channels by 2 ($r = 2$). The feature maps then pass through the KAN convolutional layer with a stride of 2 that halves the spatial dimension (to match the expected size of the feature maps at stage 3) before being expanded to 1024 channels and merge with the main network path.

The right diagram in figure shows the standard RKAN block at stage 4, which includes two KAN convolutional layers. The second layer usually processes four times the channels (\eg 2048 vs. 512) compared to the first layer, which could significantly increase the computational cost (even with degree 2 polynomials) when implemented into other previous stages.

However, our experiments on different architectures and datasets show that a second KAN layer followed by batch normalization is essential when RKAN is added to stage 4. First, features from later stages can be highly abstract, which require higher-order transformations to model these relationships between the features. The 2 KAN layers (of degree $d$ = $\{3, 2\}$) effectively creates a hierarchical degree 6 composite function, providing more expressiveness. This layer also precedes the aggregation with the main path and final classification head. Any feature degradation at this stage could inadvertently affect the final predictions. The second KAN layer allows polynomial transformations to be applied across all 2048 channels and maximizes RKAN's ability to operate on these critical high-level features.

Given that the first KAN layer $f(x)$ applies a polynomial of degree 3, where $x$ represents the input feature:
\begin{equation}
f(x) = a_3 x^3 + a_2 x^2 + a_1 x + a_0
\end{equation}

The second layer $g(y)$ applies a polynomial of degree 2, where $y = f(x)$:
\begin{equation}
g(y) = b_2 y^2 + b_1 y + b_0
\end{equation}

The composite function of degree 6 from the aggregation of 2 KAN layers, where \( b_2 (a_3 x^3)^2 = b_2 a_3^2 x^6 \), is given by:
\begin{equation}
g(f(x)) = b_2 (a_3 x^3 + a_2 x^2 + a_1 x + a_0)^2 + b_1 f(x) + b_0
\end{equation}

{
    \small
    \bibliographystyle{ieeenat_fullname}
    \bibliography{ref}

@String(CVPR= {IEEE Conf. Comput. Vis. Pattern Recog.})

@String(ICCV= {Int. Conf. Comput. Vis.})

@String(ECCV= {Eur. Conf. Comput. Vis.})

@String(ICASSP=	{ICASSP})

@String(ICLR = {Int. Conf. Learn. Represent.})

@String(CVPR  = {CVPR})

@String(ICCV  = {ICCV})

@String(ECCV  = {ECCV})

@String(ICLR  = {ICLR})

@inproceedings{simonyan2014very,
  title={Very deep convolutional networks for large-scale image recognition},
  author={Simonyan, Karen and Zisserman, Andrew},
  booktitle={International Conference on Learning Representations},
  year={2014},
  url={https://arxiv.org/abs/1409.1556}
}

@article{krizhevsky2012imagenet,
  title={ImageNet classification with deep convolutional neural networks},
  author={Krizhevsky, Alex and Sutskever, Ilya and Hinton, Geoffrey E},
  journal={Advances in neural information processing systems},
  volume={25},
  year={2012},
  url={https://papers.nips.cc/paper/2012/hash/c399862d3b9d6b76c8436e924a68c45b-Abstract.html}
}

@inproceedings{lin2014microsoft,
  title={Microsoft COCO: Common Objects in Context},
  author={Lin, Tsung-Yi and Maire, Michael and Belongie, Serge and Hays, James and Perona, Pietro and Ramanan, Deva and Doll{\'a}r, Piotr and Zitnick, C. Lawrence},
  booktitle={European Conference on Computer Vision (ECCV)},
  pages={740--755},
  year={2014},
  organization={Springer}
}

@article{khan2020survey,
  title={Survey of the recent architectures of deep convolutional neural networks},
  author={Khan, Asifullah and Sohail, Anabia and Zahoora, Umme and Qureshi, Aqsa Saeed},
  journal={Artificial Intelligence Review},
  pages={5455--5516},
  year={2020},
  publisher={Springer}
}

@article{liu2024kan,
  title={KAN 2.0: Kolmogorov-Arnold Networks Meet Science},
  author={Liu, Ziming and Ma, Pingchuan and Wang, Yixuan and Matusik, Wojciech and Tegmark, Max},
  journal={arXiv preprint arXiv:2408.10205},
  year={2024}
}

@article{lecun1989backpropagation,
  title={Backpropagation applied to handwritten zip code recognition},
  author={LeCun, Yann and Boser, Bernhard and Denker, John S and Henderson, Donnie and Howard, Richard E and Hubbard, Wayne and Jackel, Lawrence D},
  journal={Neural computation},
  volume={1},
  number={4},
  pages={541--551},
  year={1989},
  publisher={MIT Press}
}

@article{kolmogorov1957representation,
  title={On the representation of continuous functions of many variables by superposition of continuous functions of one variable and addition},
  author={Kolmogorov, Andrey Nikolaevich},
  journal={Doklady Akademii Nauk},
  volume={114},
  number={5},
  pages={953--956},
  year={1957},
  publisher={Russian Academy of Sciences}
}

@article{liu2023kolmogorov,
  title={KAN: Kolmogorov-Arnold Networks},
  author={Liu, Ziming and Wang, Yixuan and Vaidya, Sachin and Ruehle, Fabian and Halverson, James and Solja{\v{c}}i{\'c}, Marin and Hou, Thomas Y. and Tegmark, Max},
  journal={arXiv preprint arXiv:2404.19756},
  year={2024},
  url={https://arxiv.org/abs/2404.19756}
}

@article{bodner2024convolutional,
  title={Convolutional Kolmogorov-Arnold Networks},
  author={Bodner, Alexander Dylan and Tepsich, Antonio Santiago and Spolski, Jack Natan and Pourteau, Santiago},
  journal={arXiv preprint arXiv:2406.13155},
  year={2024},
  url={https://arxiv.org/abs/2406.13155}
}

@article{ss2024chebyshev,
  title={Chebyshev Polynomial-Based Kolmogorov-Arnold Networks: An Efficient Architecture for Nonlinear Function Approximation},
  author={SS, Sidharth and R, Gokul and K P, Anas and AR, Keerthana},
  journal={arXiv preprint arXiv:2405.07200},
  year={2024},
  url={https://arxiv.org/abs/2405.07200}
}

@inproceedings{he2016deep,
  title={Deep residual learning for image recognition},
  author={He, Kaiming and Zhang, Xiangyu and Ren, Shaoqing and Sun, Jian},
  booktitle={Proceedings of the IEEE conference on Computer Vision and Pattern Recognition (CVPR)},
  pages={770--778},
  year={2016},
  url={https://openaccess.thecvf.com/content_cvpr_2016/html/He_Deep_Residual_Learning_CVPR_2016_paper.html}
}

@book{goldman2002bspline,
  title={B-Spline Approximation and the de Boor Algorithm},
  author={Goldman, Ron},
  booktitle={Pyramid Algorithms},
  pages={347--443},
  year={2002},
  publisher={Elsevier},
  url={https://doi.org/10.1016/B978-155860354-7/50008-8}
}

@book{cover2012elements,
  title={Elements of information theory},
  author={Cover, Thomas M and Thomas, Joy A},
  year={2012},
  publisher={John Wiley \& Sons},
  edition={2nd}
}

@article{li2024fastkan,
  title={Kolmogorov-Arnold Networks are Radial Basis Function Networks},
  author={Li, Ziyao},
  journal={arXiv preprint arXiv:2405.06721},
  year={2024},
  url={https://arxiv.org/abs/2405.06721}
}

@book{rivlin1974chebyshev,
  title={The Chebyshev polynomials},
  author={Rivlin, Theodore J.},
  year={1974},
  publisher={John Wiley \& Sons}
}

@article{lecun2015deep,
  title={Deep learning},
  author={LeCun, Yann and Bengio, Yoshua and Hinton, Geoffrey},
  journal={Nature},
  volume={521},
  number={7553},
  pages={436--444},
  year={2015},
  publisher={Nature Publishing Group}
}

@article{lecun1998gradient,
  title={Gradient-based learning applied to document recognition},
  author={LeCun, Yann and Bottou, L{\'e}on and Bengio, Yoshua and Haffner, Patrick},
  journal={Proceedings of the IEEE},
  year={1998}
}

@inproceedings{luo2016understanding,
  title={Understanding the effective receptive field in deep convolutional neural networks},
  author={Luo, Wenjie and Li, Yujia and Urtasun, Raquel and Zemel, Richard},
  booktitle={NeurIPS},
  year={2016}
}

@book{mason2002chebyshev,
  title={Chebyshev Polynomials: Approximation Theory and Applications},
  author={Mason, John C. and Handscomb, David C.},
  year={2002},
  publisher={Chapman and Hall/CRC},
  address={Boca Raton, FL},
  isbn={978-1584882914}
}

@inproceedings{huang2017densely,
  title={Densely connected convolutional networks},
  author={Huang, Gao and Liu, Zhuang and Van Der Maaten, Laurens and Weinberger, Kilian Q.},
  booktitle={Proceedings of the IEEE Conference on Computer Vision and Pattern Recognition},
  pages={4700--4708},
  year={2017},
  url={https://openaccess.thecvf.com/content_cvpr_2017/html/Huang_Densely_Connected_Convolutional_CVPR_2017_paper.html}
}

@techreport{krizhevsky2009learning,
  title={Learning multiple layers of features from tiny images},
  author={Krizhevsky, Alex and Hinton, Geoffrey},
  institution={University of Toronto},
  year={2009}
}

@inproceedings{deng2009imagenet,
  title={Imagenet: A large-scale hierarchical image database},
  author={Deng, Jia and Dong, Wei and Socher, Richard and Li, Li-Jia and Li, Kai and Fei-Fei, Li},
  booktitle={Proceedings of the IEEE Conference on Computer Vision and Pattern Recognition (CVPR)},
  pages={248--255},
  year={2009},
  url={https://doi.org/10.1109/CVPR.2009.5206848}
}

@inproceedings{he2017mask,
  title={Mask r-cnn},
  author={He, Kaiming and Gkioxari, Georgia and Doll{\'a}r, Piotr and Girshick, Ross},
  booktitle={Proceedings of the IEEE international conference on computer vision},
  pages={2961--2969},
  year={2017}
}

@inproceedings{loshchilov2016sgdr,
  title={SGDR: Stochastic gradient descent with warm restarts},
  author={Loshchilov, Ilya and Hutter, Frank},
  booktitle={International Conference on Learning Representations (ICLR)},
  year={2017}
}

@article{goyal2017accurate,
  title={Accurate, large minibatch sgd: Training imagenet in 1 hour},
  author={Goyal, Priya and Doll{\'a}r, Piotr and Girshick, Ross and Noordhuis, Pieter and Wesolowski, Lukasz and Kyrola, Aapo and Tulloch, Andrew and Jia, Yangqing and He, Kaiming},
  journal={arXiv preprint arXiv:1706.02677},
  year={2017}
}

@inproceedings{cubuk2020randaugment,
  title={Randaugment: Practical automated data augmentation with a reduced search space},
  author={Cubuk, Ekin D and Zoph, Barret and Shlens, Jonathon and Le, Quoc V},
  booktitle={Proceedings of the IEEE/CVF Conference on Computer Vision and Pattern Recognition Workshops},
  pages={702--703},
  year={2020}
}

@inproceedings{radosavovic2020designing,
  title={Designing network design spaces},
  author={Radosavovic, Ilija and Kosaraju, Raj Prateek and Girshick, Ross and He, Kaiming and Doll{\'a}r, Piotr},
  booktitle={Proceedings of the IEEE/CVF Conference on Computer Vision and Pattern Recognition (CVPR)},
  pages={10428--10436},
  year={2020},
  url={https://openaccess.thecvf.com/content_CVPR_2020/html/Radosavovic_Designing_Network_Design_Spaces_CVPR_2020_paper.html}
}

@inproceedings{xie2017aggregated,
  title={Aggregated residual transformations for deep neural networks},
  author={Xie, Saining and Girshick, Ross and Doll{\'a}r, Piotr and Tu, Zhuowen and He, Kaiming},
  booktitle={Proceedings of the IEEE Conference on Computer Vision and Pattern Recognition (CVPR)},
  pages={1492--1500},
  year={2017},
  url={https://openaccess.thecvf.com/content_cvpr_2017/html/Xie_Aggregated_Residual_Transformations_CVPR_2017_paper.html}
}

@inproceedings{le2015tiny,
  title={Tiny ImageNet Visual Recognition Challenge},
  author={Le, Ya and Yang, Xuan S.},
  booktitle={Stanford CS 231N},
  year={2015}
}

@article{yu2024kan,
  title={KAN or MLP: A Fairer Comparison},
  author={Yu, Runpeng and Yu, Weihao and Wang, Xinchao},
  journal={arXiv preprint arXiv:2407.16674},
  year={2024},
  url={https://arxiv.org/abs/2407.16674}
}

@article{zagoruyko2016wide,
  title={Wide Residual Networks},
  author={Zagoruyko, Sergey and Komodakis, Nikos},
  journal={arXiv preprint arXiv:1605.07146},
  year={2016},
  url={https://arxiv.org/abs/1605.07146}
}

@article{hou2024kan,
  title={A Comprehensive Survey on Kolmogorov Arnold Networks (KAN)},
  author={Yuntian Hou and Di Zhang},
  journal={arXiv preprint arXiv:2407.11075},
  year={2024},
  url={https://doi.org/10.48550/arXiv.2407.11075}
}

@article{somvanshi2024survey,
  title={A Survey on Kolmogorov-Arnold Network},
  author={Somvanshi, Shriyank and Javed, Syed Aaqib and Islam, Md Monzurul and Pandit, Diwas and Das, Subasish},
  journal={arXiv preprint arXiv:2411.06078},
  year={2024},
  url={https://doi.org/10.48550/arXiv.2411.06078}
}

@inproceedings{yun2019cutmix,
  title={CutMix: Regularization Strategy to Train Strong Classifiers with Localizable Features},
  author={Yun, Sangdoo and Han, Dongyoon and Oh, Seong Joon and Chun, Sanghyuk and Choe, Junsuk and Yoo, Youngjoon},
  booktitle={Proceedings of the IEEE/CVF International Conference on Computer Vision (ICCV)},
  year={2019},
  pages={6023--6032},
  url={https://openaccess.thecvf.com/content_ICCV_2019/html/Yun_CutMix_Regularization_Strategy_to_Train_Strong_Classifiers_With_Localizable_Features_ICCV_2019_paper.html}
}

@article{keys1981cubic,
  title={Cubic Convolution Interpolation for Digital Image Processing},
  author={Keys, Robert G.},
  journal={IEEE Transactions on Acoustics, Speech, and Signal Processing},
  volume={29},
  number={6},
  pages={1153--1160},
  year={1981},
  doi={10.1109/TASSP.1981.1163711},
  url={https://ieeexplore.ieee.org/document/1163711}
}

@inproceedings{bossard14,
  title = {Food-101 -- Mining Discriminative Components with Random Forests},
  author = {Bossard, Lukas and Guillaumin, Matthieu and Van Gool, Luc},
  booktitle = {European Conference on Computer Vision},
  pages="446--461",
  year = {2014}
}

@inproceedings{sutskever2013importance,
  title={On the importance of initialization and momentum in deep learning},
  author={Sutskever, Ilya and Martens, James and Dahl, George and Hinton, Geoffrey},
  booktitle={Proceedings of the 30th International Conference on Machine Learning (ICML)},
  year={2013},
  organization={PMLR}
}

@inproceedings{zhang2017understanding,
  title={Understanding deep learning requires rethinking generalization},
  author={Zhang, Chiyuan and Bengio, Samy and Hardt, Moritz and Recht, Benjamin and Vinyals, Oriol},
  booktitle={International Conference on Learning Representations (ICLR)},
  year={2017}
}

@inproceedings{arpit2017closer,
  title={A closer look at memorization in deep networks},
  author={Arpit, Devansh and Jastrzębski, Stanisław and Ballas, Nicolas and Krueger, David and Bengio, Emmanuel and Kanwal, Maxinder S and Maharaj, Tegan and Fischer, Asja and Courville, Aaron and Bengio, Yoshua and others},
  booktitle={International conference on machine learning},
  pages={233--242},
  year={2017}
}

@inproceedings{zhang2018mixup,
  title={mixup: Beyond Empirical Risk Minimization},
  author={Zhang, Hongyi and Cisse, Moustapha and Dauphin, Yann N and Lopez-Paz, David},
  booktitle={International Conference on Learning Representations},
  year={2018}
}

@article{lin2024reducing,
  title={Reducing Data Bottlenecks in Distributed, Heterogeneous Neural Networks},
  author={Ruhai Lin and Rui-Jie Zhu and Jason K. Eshraghian},
  journal={arXiv preprint arXiv:2410.09650},
  year={2024}
}

@inproceedings{iandola2016squeezenet,
  title={SqueezeNet: AlexNet-level accuracy with 50x fewer parameters and $<$0.5 MB model size},
  author={Iandola, Forrest N and Han, Song and Moskewicz, Matthew W and Ashraf, Khalid and Dally, William J and Keutzer, Kurt},
  booktitle={International Conference on Learning Representations (ICLR)},
  year={2016}
}

@inproceedings{hu2018senet,
  author={Jie Hu and Li Shen and Gang Sun},
  title={Squeeze-and-Excitation Networks},
  booktitle={Proceedings of the IEEE Conference on Computer Vision and Pattern Recognition (CVPR)},
  year={2018},
  pages={7132--7141}
}

@inproceedings{woo2018cbam,
  title={CBAM: Convolutional Block Attention Module},
  author={Woo, Sanghyun and Park, Jongchan and Lee, Joon-Young and Kweon, In So},
  booktitle={Proceedings of the European Conference on Computer Vision (ECCV)},
  pages={3--19},
  year={2018}
}

@inproceedings{cao2019gcnet,
  title={GCNet: Non-local Networks Meet Squeeze-Excitation Networks and Beyond},
  author={Cao, Yue and Xu, Jiarui and Lin, Stephen and Wei, Fangyun and Hu, Han},
  booktitle={Proceedings of the IEEE/CVF International Conference on Computer Vision (ICCV)},
  pages={1971--1980},
  year={2019}
}

@inproceedings{yu2018deep,
    author = {Yu, Fisher and Wang, Dequan and Shelhamer, Evan and Darrell, Trevor},
    title = {Deep Layer Aggregation},
    booktitle = {Proceedings of the IEEE/CVF Conference on Computer Vision and Pattern Recognition (CVPR)},
    month = {June},
    year = {2018},
    pages = {2403--2412},
    doi = {10.1109/CVPR.2018.00255}
}

@inproceedings{he2019bag,
    author = {He, Tong and Zhang, Zhi and Zhang, Hang and Zhang, Zhongyue and Xie, Junyuan and Li, Mu},
    title = {Bag of Tricks for Image Classification with Convolutional Neural Networks},
    booktitle = {Proceedings of the IEEE/CVF Conference on Computer Vision and Pattern Recognition (CVPR)},
    month = {June},
    year = {2019},
    pages = {558--567},
    doi = {10.1109/CVPR.2019.00065}
}

@inproceedings{han2021rethinking,
    author = {Han, Dongyoon and Yun, Sangdoo and Heo, Byeongho and Yoo, YoungJoon},
    title = {Rethinking Channel Dimensions for Efficient Model Design},
    booktitle = {Proceedings of the IEEE/CVF Conference on Computer Vision and Pattern Recognition (CVPR)},
    month = {June},
    year = {2021},
    pages = {567--576}
}

@inproceedings{ridnik2021tresnet,
  title={TResNet: High performance GPU-dedicated architecture},
  author={Ridnik, Tal and Lawen, Hussam and Noy, Asaf and Ben Baruch, Emanuel and Sharir, Gilad and Friedman, Itamar},
  booktitle={Proceedings of the IEEE/CVF Winter Conference on Applications of Computer Vision},
  pages={1400--1409},
  year={2021}
}

@InProceedings{wang2020eca,
  title={ECA-Net: Efficient Channel Attention for Deep Convolutional Neural Networks},
  author={Qilong Wang and Banggu Wu and Pengfei Zhu and Peihua Li and Wangmeng Zuo and Qinghua Hu},
  booktitle={Proceedings of the IEEE/CVF Conference on Computer Vision and Pattern Recognition (CVPR)},
  pages={11534--11542},
  year={2020}
}

@inproceedings{dosovitskiy2021image,
  title={An Image is Worth 16x16 Words: Transformers for Image Recognition at Scale},
  author={Dosovitskiy, Alexey and Beyer, Lucas and Kolesnikov, Alexander and Weissenborn, Dirk and Zhai, Xiaohua and Unterthiner, Thomas and Dehghani, Mostafa and Minderer, Matthias and Heigold, Georg and Gelly, Sylvain and Uszkoreit, Jakob and Houlsby, Neil},
  booktitle={International Conference on Learning Representations (ICLR)},
  year={2021}
}

@inproceedings{liu2022convnet,
  title={A ConvNet for the 2020s},
  author={Liu, Zhuang and Mao, Hanzi and Wu, Chao-Yuan and Feichtenhofer, Christoph and Darrell, Trevor and Xie, Saining},
  booktitle={Proceedings of the IEEE/CVF Conference on Computer Vision and Pattern Recognition (CVPR)},
  pages={11976--11986},
  year={2022}
}

@article{vasu2023fastvit,
  title={FastViT: A Fast Hybrid Vision Transformer using Structural Reparameterization},
  author={Vasu, Pavan Kumar Anasosalu and Gabriel, James and Zhu, Jeff and others},
  journal={arXiv preprint arXiv:2303.14189},
  year={2023}
}

@inproceedings{tu2022maxvit,
  title={MaxViT: Multi-Axis Vision Transformer},
  author={Tu, Zhengzhong and Talebi, Hossein and Zhang, Han and Yang, Feng and Milanfar, Peyman and Bovik, Alan and Li, Yinxiao},
  booktitle={European Conference on Computer Vision (ECCV)},
  pages={459--479},
  year={2022}
}

@inproceedings{qin2024mobilenetv4,
  title={MobileNetV4: Universal Models for the Mobile Ecosystem},
  author={Qin, Danfeng and Leichner, Chas and Delakis, Manolis and Fornoni, Marco and Luo, Shixin and Yang, Fan and Wang, Weijun and Banbury, Colby and Ye, Chengxi and Akin, Berkin and Massa, Vaibhav and Narayanan, Andrey and Howard, Andrew},
  booktitle={European Conference on Computer Vision (ECCV)},
  year={2024}
}

@inproceedings{tan2021efficientnetv2,
  title={EfficientNetV2: Smaller Models and Faster Training},
  author={Tan, Mingxing and Le, Quoc},
  booktitle={ICML},
  year={2021}
}

@inproceedings{dai2021coatnet,
  title={CoAtNet: Marrying Convolution and Attention for All Data Sizes},
  author={Dai, Zihang and Liu, Hanxiao and Le, Quoc V and Tan, Mingxing},
  booktitle={NeurIPS},
  year={2021}
}

@article{zhang2020resnest,
  title={ResNeSt: Split-Attention Networks},
  author={Zhang, Hang and Wu, Chongruo and Zhang, Zhongyue and others},
  journal={arXiv preprint arXiv:2004.08955},
  year={2020}
}

@article{gao2019res2net,
  title={Res2Net: A New Multi-scale Backbone Architecture},
  author={Gao, Shang-Hua and Cheng, Ming-Ming and Zhao, Kai and others},
  journal={IEEE TPAMI},
  year={2019}
}

@inproceedings{yang2021simam,
  title={SimAM: A Simple, Parameter-Free Attention Module for Convolutional Neural Networks},
  author={Yang, Lingxiao and Zhang, Ru-Yuan and Li, Lida and Xie, Xiaohua},
  booktitle={International Conference on Machine Learning (ICML)},
  pages={11863--11874},
  year={2021},
  organization={PMLR}
}

@article{xu2024ela,
  title={ELA: Efficient Local Attention for Deep Convolutional Neural Networks},
  author={Xu, Wei and Wan, Yi},
  journal={arXiv preprint arXiv:2403.01123},
  year={2024}
}

@inproceedings{zhang2021sa,
  title={SA-Net: Shuffle Attention for Deep Convolutional Neural Networks},
  author={Zhang, Qing-Long and Yang, Yu-Bin},
  booktitle={2021 IEEE International Conference on Acoustics, Speech and Signal Processing (ICASSP)},
  pages={2235--2239},
  year={2021},
  organization={IEEE}
}

@inproceedings{han2017deep,
  title={Deep pyramidal residual networks},
  author={Han, Dongyoon and Kim, Jiwhan and Kim, Junmo},
  booktitle={Proceedings of the IEEE conference on computer vision and pattern recognition},
  pages={5927--5935},
  year={2017}
}

@inproceedings{liu2023efficientvit,
  title={EfficientViT: Memory Efficient Vision Transformer with Cascaded Group Attention},
  author={Liu, Xinyu and Peng, Houwen and Zheng, Ningxin and others},
  booktitle={CVPR},
  year={2023}
}

@inproceedings{lin2017fpn,
  title={Feature Pyramid Networks for Object Detection},
  author={Lin, Tsung-Yi and Doll{\'a}r, Piotr and Girshick, Ross and He, Kaiming and Hariharan, Bharath and Belongie, Serge},
  booktitle={Proceedings of the IEEE Conference on Computer Vision and Pattern Recognition (CVPR)},
  year={2017}
}

@inproceedings{liu2018pan,
  title={Path Aggregation Network for Instance Segmentation},
  author={Liu, Shu and Qi, Lu and Qin, Haifang and Shi, Jianping and Jia, Jiaya},
  booktitle={Proceedings of the IEEE Conference on Computer Vision and Pattern Recognition (CVPR)},
  year={2018}
}
}

% WARNING: do not forget to delete the supplementary pages from your submission 
% \input{sec/X_suppl}

\end{document}